\documentclass[a4paper,fleqn]{cas-dc}

\usepackage[authoryear]{natbib}
\usepackage{indentfirst}
\makeatletter
\patchcmd{\paragraph}{\parindent}{\paraindent}{}{%
  \PackageError{SAR}{Could not set paragraph heading indentation}{}
}
\patchcmd{\paragraph}{\itshape}{\upshape\bfseries}{}{%
  \PackageError{SAR}{Could not set paragraph heading font}{}
}
\apptocmd{\@enum@}{%
  \setlength{\itemindent}{\paraindent}%
  \setlength{\listparindent}{\paraindent}%
  \setlength{\parindent}{\paraindent}%
}{}{\PackageError{SAR}{Could not set numbered-list indentation}{}}
\apptocmd{\@Itemize}{%
  \setlength{\itemindent}{\paraindent}%
  \setlength{\listparindent}{\paraindent}%
  \setlength{\parindent}{\paraindent}%
}{}{\PackageError{SAR}{Could not set bulleted-list indentation}{}}
\renewcommand{\@seccntformat}[1]{%
  \csname the#1\endcsname%
  \ifstrequal{#1}{subsection}{}{\@seccntDot}%
  \hskip 0.5em}
\makeatother

\ExplSyntaxOn
\cs_set:Npn \__first_footerline:
  {
    \group_begin:
    \small
    \sffamily
    \ifnum\theblind>0\relax
    \else
      \__short_authors: :~
    \fi
    { \rmfamily \itshape Preprint }
    \group_end:
  }
\ExplSyntaxOff

\def\tsc#1{\csdef{#1}{\textsc{\lowercase{#1}}\xspace}}
\tsc{WGM}
\tsc{QE}

\begin{document}
\let\WriteBookmarks\relax
\def\floatpagepagefraction{1}
\def\textpagefraction{.001}

\shorttitle{}
\shortauthors{Xinpeng Lei et~al.}

\title[mode=title]{SAR: Learning Global Item--Space Matching and Geometry-Aware Placement Ranking for Buffered Online 3D Bin Packing}

\author[1]{Xinpeng Lei}[orcid=0009-0002-2916-1601]
\ead{xinpeng.lei@stu.xbmu.edu.cn}
\author[2]{Jin Zhang}[orcid=0009-0000-9325-8251]
\ead{zhangjin2024@lzu.edu.cn}
\author[2]{Yizhe Yang}[orcid=0009-0007-2573-0041]
\ead{yyizhe2023@lzu.edu.cn}
\author[2]{Yihong Duan}[orcid=0009-0003-0700-0505]
\ead{duanyh2024@lzu.edu.cn}
\author[1]{Fucheng Wan}[orcid=0009-0007-6084-831X]
\cormark[1]
\ead{wanfucheng@126.com}

\affiliation[1]{organization={Northwest Minzu University},
                city={Lanzhou},
                state={Gansu},
                country={China}}

\affiliation[2]{organization={Lanzhou University},
                city={Lanzhou},
                state={Gansu},
                country={China}}

\cortext[cor1]{Corresponding author}

\begin{abstract}
Online 3D bin packing with a bounded buffer requires irreversible item-selection and placement decisions under limited knowledge of future arrivals.
Feasibility alone is insufficient, as feasible actions can produce different residual-space structures and subsequent packing opportunities.
A central architectural question is how to allocate spatial reasoning between item selection and placement evaluation.
We propose Spatial Affordance Retrieval (SAR), a novel learning-based framework that coordinates global item--space matching with geometry-aware placement ranking.
A voxel-token spatial encoder represents 3D container occupancy, enabling a picker to match buffered items to the residual space without requiring item-specific placement proposals as inputs.
After item selection, a loader ranks feasible rotation--position candidates using candidate geometry, local geometric features (LGF), and inter-candidate comparison, without directly accessing global spatial tokens.
The feasible candidate set connects the two stages, which are jointly optimized under a shared long-horizon packing return.
On the Random, JD, and SF datasets, the best SAR variants improve utilization over the strongest evaluated online baselines by 0.96--4.20 percentage points under the no-buffer setting ($K=1$).
With a buffer ($K=5$), the best utilization rates across SAR variants reach 93.55\%, 91.04\%, and 90.41\%, respectively.
Ablations support the complementary roles of global item--space matching and geometry-aware placement evaluation, counterfactual analysis suggests that learned item selection favors greater future packing potential, and experiments demonstrate compatibility with multiple candidate generators.
\end{abstract}


\maketitle
\hypersetup{pdfauthor={Xinpeng Lei, Jin Zhang, Yizhe Yang, Yihong Duan, Fucheng Wan}}

\section{Introduction}

\begin{figure}[pos=!t]
\centering
\includegraphics[width=\columnwidth]{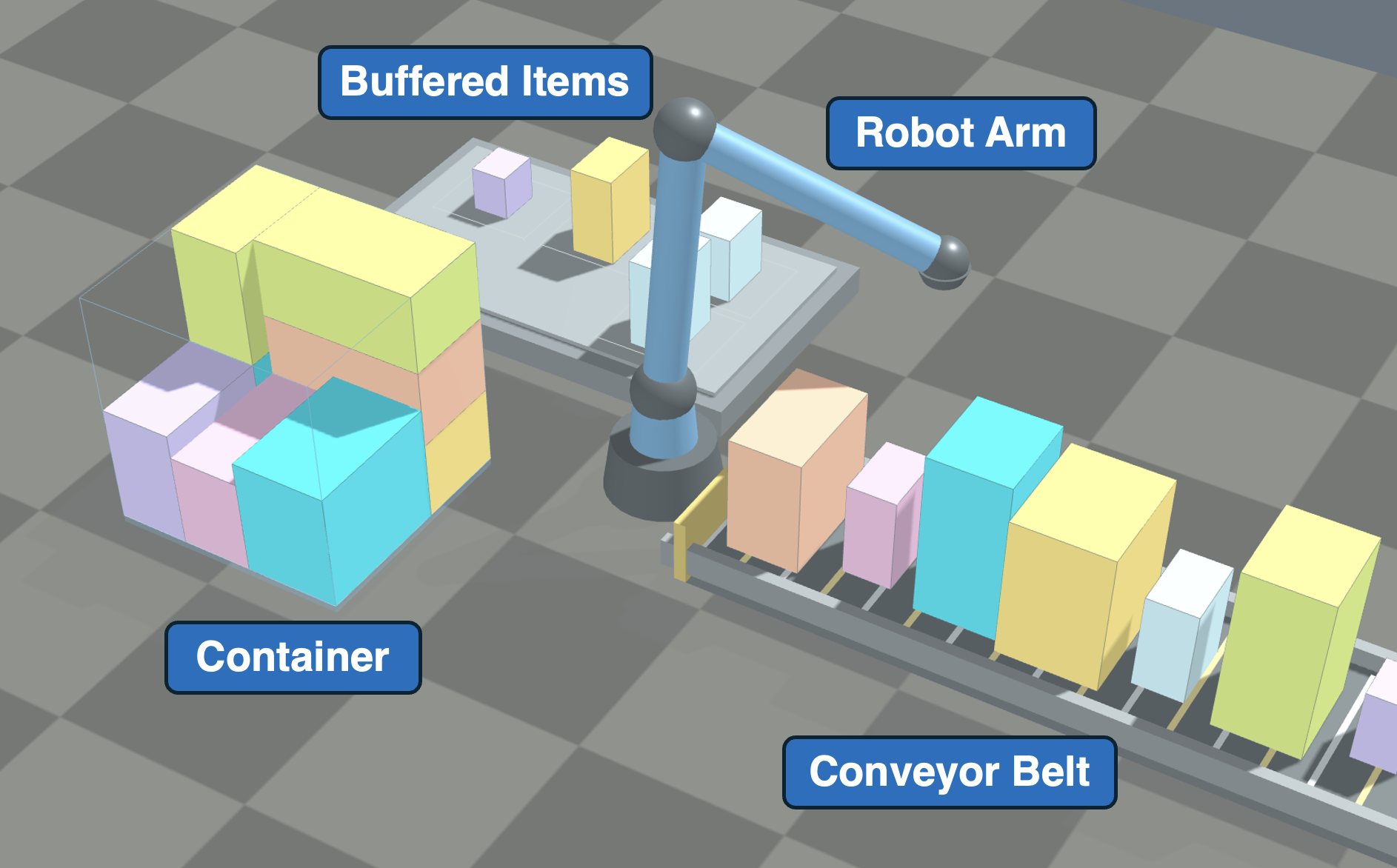}
\caption{Illustration of the buffered online 3D-BP scenario.  Items arrive on a conveyor and enter a limited buffer. The policy selects which item to pack and its orientation and position in the container. The robot arm visualizes the selected item's transfer and placement.}
\label{fig:simulation-environment}
\end{figure}

The rapid growth of e-commerce and modern logistics has placed increasing demand on automated warehousing and robotic packing, where efficiently loading items into containers directly affects transportation cost, storage footprint, and throughput \citep{dahmani2025reinforcement}. 
At the core of these systems lies the online 3D bin packing (3D-BP), which requires sequentially placing cuboid items into a container to maximize space utilization under limited knowledge of future arrivals \citep{zhao2021online:8}. 
The problem is NP-hard \citep{cp}, and its online, irreversible nature makes it especially challenging: each placement is committed immediately and reshapes the residual space, constraining all subsequent packing opportunities. 
In the bounded-buffer setting considered in this work, the policy must therefore jointly decide which visible item to pack and where to place it.
Figure~\ref{fig:simulation-environment} illustrates the simulated experimental
scenario.

Existing 3D-BP methods can be broadly divided into heuristic and learning-based approaches \citep{wu2023machine}. Heuristic methods are efficient but depend on manually designed rules, whereas recent deep reinforcement learning (DRL) methods learn adaptive policies for sequential packing and have shown strong performance in complex settings \citep{dahmani2025reinforcement}. Yet in crowded containers, feasibility alone is not enough: different feasible actions can induce markedly different residual-space structures and hence different future packing potential. Early deep learning methods predicted joint packing actions in a single step, but learning difficulties and performance bottlenecks have motivated hierarchical networks based on action decomposition, whose effectiveness has been demonstrated despite the lack of clear principles for decomposing packing decisions. 

\begin{figure}
\centering
\includegraphics[width=0.99\columnwidth]{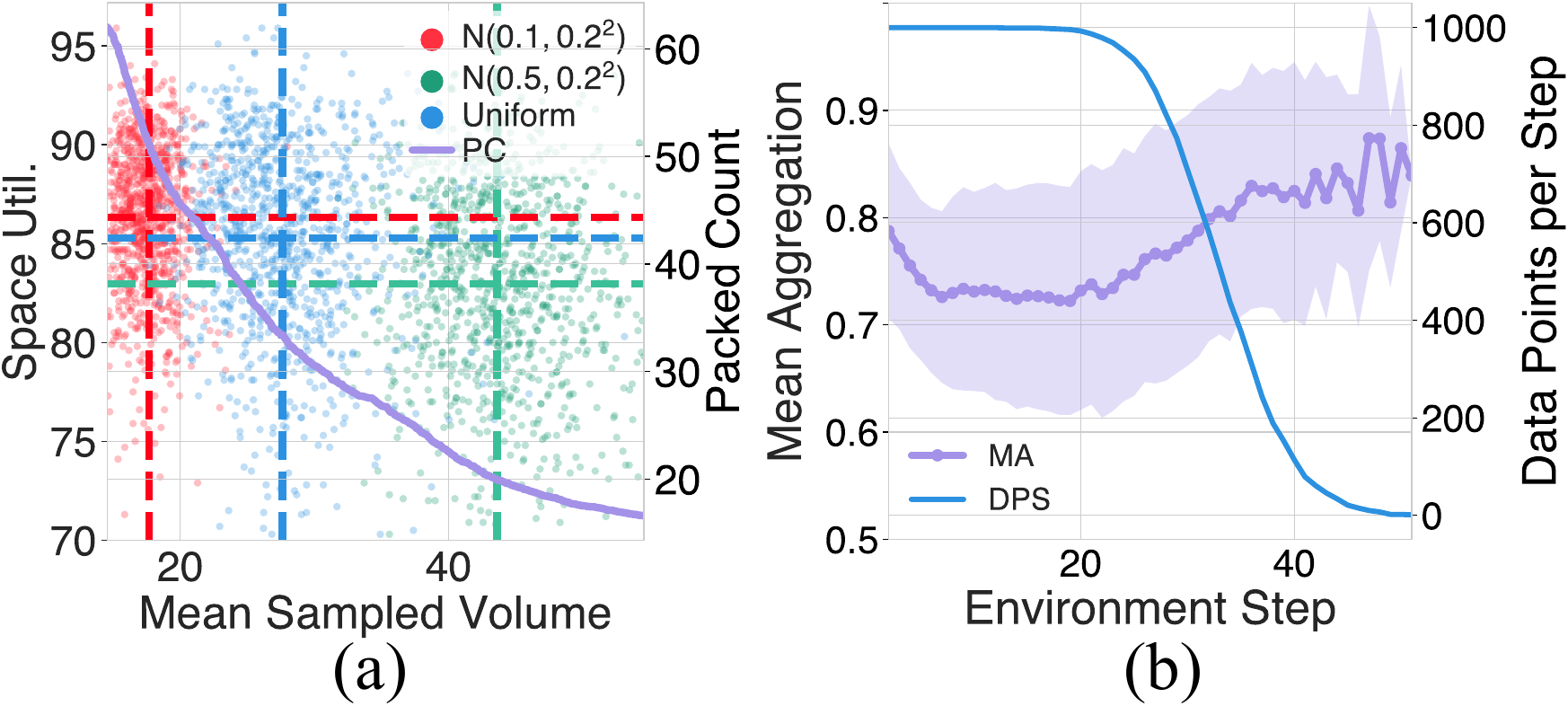} %
\caption{Empirical observations. (a)  Across item-size distributions, smaller items are associated with higher packed-item counts (longer packing trajectories) and higher final utilization, whereas larger items exhibit the opposite pattern. (b) As packing progresses, feasible placement candidates become fewer and more spatially aggregated, increasing the need for fine-grained candidate evaluation.}
\label{fig:motivation}
\end{figure}

To identify performance bottlenecks and opportunities for improvement, we conducted empirical analyses based on existing methods. \textbf{First}, Figure~\ref{fig:motivation}(a) shows that distributions with smaller items yield more packed items and hence longer packing trajectories. These distributions also achieve higher final utilization, while distributions with larger items exhibit the opposite pattern. Although item size jointly affects both quantities, their co-variation suggests that sustaining feasible placements is associated with more complete exploitation of the residual space. \textbf{Second}, Figure~\ref{fig:motivation}(b) shows that feasible placement candidates become fewer and more spatially aggregated as packing progresses. Although these candidates satisfy the same feasibility constraints, they can still differ in their effects on local space structure. Together, these observations motivate evaluating item choices against the global residual space and distinguishing placement candidates by their local geometric effects, particularly in high-occupancy states.

Existing work has explored buffered item selection \citep{sarawgi2026preference}, fine-grained placement quality evaluation \citep{poolavaram2026operationally}, and joint item--placement learning through hierarchical policies \citep{wang2026robotic}.
These advances motivate a central architectural question: how should spatial reasoning be allocated between item selection and placement in buffered online packing?
Two coupled design choices follow: how to represent the global residual space for matching buffered items, and which geometric cues and contextual interactions are needed to rank a selected item's feasible placements.
Some placement-aware designs combine several relational and contextual components whose individual contributions are not fully disentangled, motivating a focused examination of the interactions required for placement ranking.

With a volume-based per-step reward, all feasible placements of a selected item yield the same immediate gain.
Item selection therefore affects both immediate gain and future reward, whereas feasible rotation--position choices for that item affect only future reward.
This distinction motivates assigning global item--space matching to item selection and local geometric comparison to placement ranking.

We propose \textbf{Spatial Affordance Retrieval (SAR)}, a novel learning-based framework for online 3D-BP.
SAR adopts a modular design with the feasible candidate set as an explicit interface between item selection and placement evaluation.
A \textbf{Spatial Encoder} converts binary 3D voxel occupancy into spatial tokens, which the \textbf{picker} uses to match buffered items to the global residual-space structure.
For the selected item, candidate generation and feature computation produce feasible rotation--position candidates with Local Geometric Features (LGF).
The \textbf{loader} performs fine-grained evaluation of this set using candidate geometry, LGF, and inter-candidate comparison, without directly accessing global spatial tokens.
The two stages are jointly optimized under a shared long-horizon packing return.
Here, \emph{spatial affordance} denotes learned, state-conditioned suitability for exploiting residual space, and \emph{retrieval} denotes scoring and selecting among available items and feasible placements.

Our main contributions are summarized as follows:

\begin{enumerate}[1.]
    \setlength{\itemindent}{0pt}
    \setlength{\listparindent}{0pt}
    \setlength{\parindent}{0pt}
    \item \textbf{3D occupancy-conditioned item--space matching.}
    We develop a voxel-token spatial encoder that extracts local occupancy geometry and global spatial context for matching buffered items to the residual space.
    The picker conditions item selection on this representation without requiring item-specific placement proposals as inputs, while detailed rotation--position evaluation is deferred to the loader.

    \item \textbf{Spatial Affordance Retrieval.}
    We propose SAR, a novel learning-based framework for buffered online 3D-BP, whose voxel-based picker and LGF-based loader are jointly designed for item selection and fine-grained placement evaluation.
    Both stages are optimized under a shared packing objective.

    \item \textbf{Experimental validation.}
    Experiments across three datasets and multiple candidate generators demonstrate improved utilization.
    Ablations support the complementary roles of the two stages, while counterfactual analysis suggests that learned item selection favors greater future packing potential.
\end{enumerate}

\section{Related Work}

Research on 3D-BP spans geometric heuristics, search-based optimization, and learned sequential policies.
Beyond this broad classification, methods differ in the items they can observe, the decisions they learn, and the spatial information used to evaluate those decisions.
We organize the discussion around candidate generation, action decomposition, spatial representation, and placement quality evaluation.
Table~\ref{tab:related-comparison} summarizes representative designs, distinguishing local from global item visibility and fixed from adjustable item order, alongside learned item and placement decisions.

\begin{table*}[t]
\caption{Comparison of representative packing policies. Item visibility is global when the policy observes the complete item set and local when it observes only the current item or a bounded buffer. Item order is fixed or adjustable during packing. Learned decisions are listed in the order item, rotation, and position, where a parenthesized group denotes joint selection and a comma between groups denotes separate stages; decisions not learned by the policy itself are omitted. Entries refer to the formulations discussed in the text.}
\label{tab:related-comparison}
\centering
\scriptsize
\setlength{\tabcolsep}{3pt}
\renewcommand{\arraystretch}{1.18}
\begin{tabular*}{\textwidth}{@{\extracolsep{\fill}}>{\raggedright\arraybackslash}p{0.11\textwidth}>{\raggedright\arraybackslash}p{0.055\textwidth}>{\raggedright\arraybackslash}p{0.065\textwidth}>{\raggedright\arraybackslash}p{0.08\textwidth}>{\raggedright\arraybackslash}p{0.15\textwidth}>{\raggedright\arraybackslash}p{0.12\textwidth}>{\raggedright\arraybackslash}p{0.175\textwidth}>{\raggedright\arraybackslash}p{0.125\textwidth}@{}}
\toprule
Method & Type & Item visibility & Item order & Learned decisions & Spatial representation & Placement evaluation & Training \\
\midrule
PackIt \citep{goyal2020packit} & Offline & Global & Adjustable & Item & 3D voxel grid & Handcrafted; BLBF placement with Aligned rotation & PPO \\
\addlinespace
TAP-Net \citep{hu2020tap} & Offline & Global & Adjustable & \mbox{(Item, rotation)} & 2D height map & Handcrafted criteria inside the heuristic; no learned evaluation & Policy gradient; heuristic packing \\
\addlinespace
Attend2Pack \citep{zhang2021attend2pack} & Offline & Global & Adjustable & Item, \mbox{(rotation, position)} & 2D frontier map & Implicit; from oriented dimensions and frontier geometry & REINFORCE with rollout baseline \\
\addlinespace
PCT \citep{pct:1} & Online & Local & Fixed & \mbox{(Rotation, position)} & Tree structure & Implicit; from placed-item and candidate-node geometry & ACKTR \\
\addlinespace
Zhao et al. \citep{zhao2022learning:2} & Online & Local & Fixed & Rotation, position & 2D height map & Implicit; from height-map geometry; stacking tree for feasibility & Constrained ACKTR \\
\addlinespace
GOPT \citep{gopt:9} & Online & Local & Fixed & \mbox{(Rotation, position)} & 2D height map & Implicit; from free-space and item--space relations & PPO \\
\addlinespace
STEP \citep{sarawgi2026preference} & Online & Local & Adjustable & Item & 2D height map & External; supplied by a separate GOPT policy & Preference-conditioned RL \\
\addlinespace
\citet{wang2026robotic} & Online & Local & Adjustable & Item, \mbox{(rotation, position)} & Tree structure & Implicit; mixed attention over placed-item and candidate geometry & Dueling DQN \\
\addlinespace
OPAL \citep{poolavaram2026operationally} & Online & Local & Fixed & \mbox{(Rotation, position)} & 2D height map & Explicit; geometric and operational descriptors exposed per candidate & PPO \\
\addlinespace
\textbf{SAR (ours)} & \textbf{Online} & \textbf{Local} & \textbf{Adjustable} & \textbf{Item, \mbox{(rotation, position)}} & \textbf{3D voxel grid} & \textbf{Explicit; geometric descriptors (LGF) exposed per candidate} & \textbf{PPO} \\
\bottomrule
\end{tabular*}
\end{table*}

\subsection{Geometric Heuristics and Candidate Generation}

Geometric heuristics reduce the placement search space through structured candidate construction.
Corner Point (CP) methods identify candidate anchors from the packing configuration \citep{cp}, while Extreme Point (EP) methods construct and update points induced by placed items \citep{ep}.
Empty Maximal Space (EMS) methods describe available rectangular regions and derive placements within them \citep{ha2017online}.
Event Point (EV) generation provides another way to discretize placements from the evolving configuration \citep{pct:1}.
These schemes offer different candidate coverage and computational costs; physical constraints, such as support and stability, additionally determine which generated placements are admissible.

Candidate construction can be combined with item ordering, placement scores, and search over packing sequences.
Classical approaches include hybrid genetic algorithms, layer-based construction with tabu search, guided local search, and beam search \citep{bortfeldt2001hybrid,lodi2002,faroe2003guided,araya2014beam}.
Recent industrial optimization also remains relevant: GENPACK combines a genetic search with domain-specific construction procedures and multiple operational criteria \citep{poolavaram2026genpack}.
These search procedures illustrate how geometric construction and operational preferences jointly shape a packing solution, and provide useful baselines for learned policies.

For learned policies, candidate generation and candidate selection serve different purposes.
Generation determines which actions are available; selection determines which feasible action is preferable for the packing objective.
Heuristics already encode geometric preferences through anchor construction, pruning, or handcrafted scores.
Nevertheless, exposing only a candidate's coordinates does not explicitly describe its support, contacts, or fit to the surrounding geometry.
SAR retains heuristic generation and physical validation, while making such local geometric cues available to the learned placement evaluator.

\subsection{Learning-based Packing Policies and Action Decomposition}

Item selection, rotation selection, and position selection need not be learned through the same action structure.
In flexible offline packing, \citet{duan2018multi} use a shared representation with separate sequence and orientation learners, while determining placement through a heuristic.
TAP-Net instead learns an item--orientation choice and uses a packing heuristic to determine its position \citep{hu2020tap}.
Its transport-and-pack setting additionally constrains which items can be extracted from the initial arrangement.
For large-container offline packing, \citet{yin2024solving} couple item and orientation selection before learning position selection, using bidirectional packing to constrain the position search.
Their subsequent framework further develops this two-stage bidirectional design \citep{offline}.
Attend2Pack learns item selection followed by joint orientation and placement-coordinate prediction, with the remaining coordinates obtained through a geometric rule \citep{zhang2021attend2pack}.
Its main formulation observes the complete item set and trains both policies using a terminal packing objective.
These designs combine learned decisions with geometric procedures, assigning different roles to the three packing components.

For fixed-arrival online packing, \citet{zhao2022learning:2} instead factorize placement into conditional coordinate and orientation heads, without learning which item to process next.
Other fixed-order methods learn orientation and position separately, combining item reorientation with a physics-based stability heuristic \citep{zhou2024efficient}.
Thus, similar component names can describe different decision scopes and conditioning structures.

Information availability is equally important when interpreting selection policies.
Offline methods can reorder a complete set; fixed-arrival online methods must place the current item; buffered or semi-online methods can select among a limited visible set.
Adjustable-order semi-online packing has explicitly been studied by \citet{yin2025adjustable}.
STEP learns preference-conditioned item--grasp-face selection from a buffer, while obtaining placement and rotation suggestions from an external GOPT policy \citep{sarawgi2026preference}.
This differs from learning both selection and placement under a shared objective.
For buffered robotic packing, \citet{wang2026robotic} learn item selection followed by joint rotation--position selection under a shared stepwise volume reward.
Lookahead search is also distinct from actual reordering: \citet{fang2026effective} use visible future parcels to improve placement planning while preserving the arrival order.
These distinctions prevent attributing gains from a larger information horizon solely to network architecture.

Other online formulations expand the available decisions: \citet{song2023towards} learn packing and unpacking to revise earlier placements, while \citet{tsang2025deep} address concurrent dual-bin packing with bin-replacement strategies.

SAR addresses irreversible packing within a single container and adopts item selection followed by joint rotation--position selection, a decomposition also used in buffered robotic packing \citep{wang2026robotic}.
It organizes the two stages around voxel-based item--space matching and placement evaluation with explicit local geometric features.
With a volume-based per-step reward, item selection affects both immediate and future rewards, whereas feasible rotation--position choices for the selected item affect only future reward.

\subsection{Spatial Representations and Item--Space Matching}

The spatial representation determines what a policy can infer about the remaining container geometry.
Height-map policies summarize the upper occupied surface on a two-dimensional grid and support efficient convolutional encoding and feasibility prediction \citep{zhao2021online:8,zhao2022learning:2}.
For variable-height packing, \citet{que2023solving} augment height maps with plane features describing horizontal clearances and encode the state with a Transformer policy.
This is compact, but the representation alone does not distinguish all occupancy configurations beneath that surface.
Its efficiency therefore comes with a trade-off in the volumetric details retained for subsequent decisions.
For buffered selection, these details matter because several visible items may fit the same surface configuration in different ways.

Structured representations offer alternatives to dense surface maps.
Packing Configuration Trees (PCT) represent placed items as internal nodes and candidate rotation--position actions as leaf nodes, supporting attention-based evaluation and selection of candidate placements \citep{pct:1}.
PCT encodes spatial relations among configuration nodes using graph attention networks (GATs) \citep{gat}.
\citet{wang2026robotic} build on this tree representation for buffered packing: a manager selects an item from the buffer, and a worker selects an EMS-generated placement leaf.
Their networks use mixed self-attention over geometric node embeddings and are trained with separate Dueling DQNs under a shared stepwise volume reward.
Both methods encode the packing configuration through discrete item and candidate nodes.
Although these nodes contain three-dimensional coordinates, the tree representation does not directly supply a unified volumetric occupancy field to the policy; relationships among occupied regions and residual cavities must be inferred from node geometry.

GOPT represents free spaces through EMS tokens and uses attention to model both relations among spaces and their compatibility with the current item \citep{gopt:9}.
STEP extends EMS-based reasoning to a buffer of item--face options \citep{sarawgi2026preference}.
These methods already demonstrate the value of relational spatial reasoning.
For buffered packing, spatial reasoning additionally needs to compare the compatibility of several visible items with the residual space before committing to a placement.

Voxel representations explicitly describe occupied and empty regions at a chosen spatial resolution.
Prior work has demonstrated their usefulness for explicit geometric optimization in irregular packing \citep{lamas2023voxel}.
PackIt learns item selection by concatenating candidate-item, max-pooled item-set, and container features and applying nonlinear MLP scoring \citep{goyal2020packit}.

These considerations motivate using explicit 3D occupancy to match buffered items to the global residual space.
SAR learns item--space matching from a 3D spatial representation of container occupancy by separately projecting item and spatial features and averaging their scaled dot-product compatibility scores.
This scoring introduces an explicit multiplicative interaction between item and spatial representations for buffered item selection.

\begin{figure*}[pos=!t]
  \centering
  \includegraphics[width=0.99\textwidth]{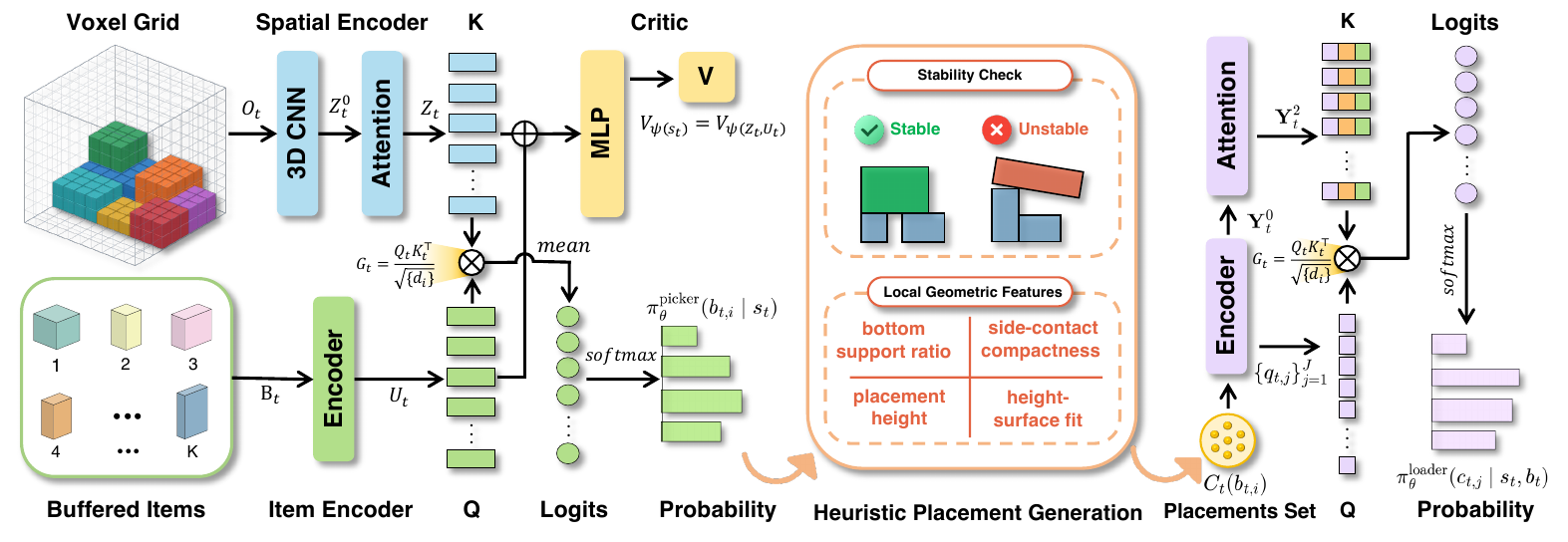} %
  \caption{Overview of the proposed Spatial Affordance Retrieval (SAR) framework. A 3D CNN with max pooling and self-attention encodes the container voxel grid $\mathbf{O}_t$ (top left) into spatial tokens $\mathbf{Z}_t$, and an item encoder embeds the buffered items $B_t$ (bottom left) into $\mathbf{U}_t$. The picker computes scaled dot-product scores between item embeddings (Q) and these tokens (K), averages the scores, and softmax-normalizes them into the item-selection policy $\pi^{\mathrm{picker}}_{\theta}(b_{t,i}\mid s_t)$; a critic additionally estimates the state value $V_{\psi}(s_t)$ for training. Given the selected item, heuristic placement generation enumerates rotation--position candidates, filters them by stability, and computes Local Geometric Features (LGF; support, compactness, height, and surface fit), forming the feasible placement set $C_t(b_{t,i})$ (middle). For fine-grained placement evaluation, the loader encodes these candidates into tokens $\mathbf{Y}^{0}_t$, refines them by self-attention into $\mathbf{Y}^{2}_t$, and scores them by scaled dot-products between placement queries $\{q_{t,j}\}_{j=1}^{J}$ (Q) and the refined keys (K), yielding the placement policy $\pi^{\mathrm{loader}}_{\theta}(c_{t,j}\mid s_t,b_t)$ that determines the final orientation and position (right). The selected item determines the loader's feasible candidate set, and both stages are trained under a shared return.}
  \label{fig2}
\end{figure*}

\subsection{Placement Quality Evaluation and Candidate Ranking}

Feasibility masks exclude invalid actions but do not rank the remaining placements by their longer-term consequences.
Policies using height maps, configuration trees, or free-space tokens can learn these preferences implicitly \citep{zhao2021online:8,pct:1,gopt:9}.
Explicit placement descriptors offer a complementary route by exposing geometric or operational properties before policy scoring.
However, feature selection and relational modeling are separate design choices: supplying local quality cues does not by itself establish which interactions should be learned, or how much additional global context a placement-ranking network requires.

Handcrafted placement evaluation also remains relevant: \citet{wang2021dense} use a heightmap-minimization heuristic for irregular-object packing subject to stability and robot-motion feasibility.

OPAL demonstrates the effectiveness of placement-aware learned evaluation in industrial packing \citep{poolavaram2026operationally}.
It enriches EMS-derived candidates with geometric and operational descriptors, encodes dependencies among attributes within each candidate, and mixes candidate representations with oriented-item and pallet-state representations.
Its evaluated workflow uses a prescribed item order, while the policy selects among feasible rotation--position candidates.
Its configuration-level comparisons do not fully isolate the contributions of every relational component, motivating further analysis of the interactions needed for effective placement evaluation.

Like OPAL, SAR ranks generated placement candidates; the architectural comparison concerns the scope of feature encoding and contextual interaction.
SAR focuses the loader on the selected item's feasible candidate set.
Its four LGF expose bottom support, side-contact compactness, placement height, and height-surface fit, supplementing coordinates and rotated item dimensions.
These features convey state-derived local context, enabling comparison among candidate tokens without an additional learned global container-context encoder at this stage.
Global item--space matching remains the picker's responsibility, and both stages are optimized using a shared packing return.
SAR thus coordinates voxel-based item matching and focused local placement ranking through a clear division of decision roles, with shared training connecting their effects on residual-space utilization.

\section{Method}

Figure~\ref{fig2} presents the SAR framework, which combines voxel-based item--space matching with local ranking of feasible placement candidates.

\subsection{Voxelized Environment Representation}

\paragraph{Voxel Representation.}
The container is discretized into a regular voxel grid $\Omega_R=\{0,\ldots,R-1\}^3,$ $R$ denotes resolution. Each voxel $q=(x,y,z)\in\Omega_R$ defines both the minimum perceptual unit of the environment and a discrete anchor for placement execution. At time step $t$, the container state is represented as a single-channel occupancy grid $\mathbf{O}_t:\Omega_R\rightarrow\{0,1\}, \mathbf{O}_t\in\{0,1\}^{R\times R\times R}.$
The occupancy value $\mathbf{O}_t(q)$ equals 1 when voxel $q$ is occupied and 0 when $q$ is empty. The picker is given a candidate item buffer $B_t = \{b_{t,1}, b_{t,2}, \ldots, b_{t,K}\}$, $K$ denotes buffer size. Each candidate item is represented by its dimensions and mass, $b_{t,i} = (d_{x,i}, d_{y,i}, d_{z,i}, m_i)$, where $m_i$ is used for stability checks. A binary mask $\mathbf{M}_t^{\mathrm{buf}} \in \{0,1\}^K$ indicates valid candidates. The complete observation is therefore written as $s_t = (\mathbf{O}_t, B_t, \mathbf{M}_t^{\mathrm{buf}})$. $\mathbf{O}_t$ is updated after $b_{t,i}$ is successfully placed. This voxelized container represents the placed items as a unified occupied solid rather than as a set of mutually dependent stacked boxes. Thus, the policy does not need to explicitly infer stacking dependencies among placed items, but can instead focus on global spatial state of the container.

\paragraph{Heuristic Placement Generation.}
SAR supports various placement generation schemes, including CP, EP, EMS, EV, and full coordinate (FC). Candidates are incrementally generated based on the newly placed item. Following \citep{zhao2022learning:2}, stability checking is enabled in all experiments to filter these candidates into a feasible placement set $\mathcal{C}_t$, separating physical validation from decision making. An effective generation scheme reduces the search space while preserving feasible packing options.

\paragraph{Local Geometric Features.}
For each feasible candidate set $\mathcal{C}_t=\{c_{t,1},c_{t,2},...,c_{t,j}\}$, the environment directly computes LGF. For each feasible placement candidate $c_t\in\mathcal{C}_t$, we denote these environment-computed features as 
\begin{equation}
\label{eq:lgf-definition}
g_{\mathrm{env}}(c_t;\mathbf{O}_t) = (\rho_{\mathrm{sup}}(c_t), \rho_{\mathrm{comp}}(c_t), h_{\mathrm{norm}}(c_t), \rho_{\mathrm{fit}}(c_t)),
\end{equation}
 where $\rho_{\mathrm{sup}}$ measures bottom support ratio, $\rho_{\mathrm{comp}}$ measures side-contact compactness, $h_{\mathrm{norm}}$ is the normalized placement height, and $\rho_{\mathrm{fit}}$ measures height-surface fit under the candidate footprint. LGF does not directly define packing quality. Instead, the policy learns how these local cues relate to the long-term packing return.

For a candidate item with anchor $p=(x,y,z)$ and rotated dimensions $(d_x,d_y,d_z)$ on the voxel grid, let $O(u,v,w)\in\{0,1\}$ denote the current occupancy grid and let $H(u,v)$ be the height map. The footprint of the candidate is:
\begin{equation}
  \mathcal{F}(p)=
  \{(u,v)\mid x \le u < x+d_x,\; y \le v < y+d_y\}.
\end{equation}

$\rho_{\mathrm{sup}}$ is computed from the occupied voxels immediately below the footprint, where $\chi_{uv}$ is a binary support indicator for footprint cell $(u,v)$.
\begin{align}
  \rho_{\mathrm{sup}}
  &=
  \begin{cases}
    1, & z=0,\\
    \frac{1}{|\mathcal{F}(p)|}
    \sum\limits_{(u,v)\in \mathcal{F}(p)} \chi_{uv}, & z>0,
  \end{cases} \\
  \chi_{uv}
  &= \mathbf{1}\{O(u,v,z-1)>0\}.
\end{align}

$\rho_{\mathrm{comp}}$ measures contact with neighboring boxes and container walls using only the four vertical side faces. Let $S_{\mathrm{side}}=2(d_y d_z+d_x d_z)$ denote the total side area. $A_{\mathrm{box}}$ is computed by counting occupied voxels in the one-voxel-thick layer adjacent to each side face that is not on a container boundary. $A_{\mathrm{wall}}$ is computed by adding the full area of each side face that coincides with a container wall, i.e., faces at $x=0$ or $x+d_x=R$ contribute $d_y d_z$, and faces at $y=0$ or $y+d_y=R$ contribute $d_x d_z$. The compactness ratio is
\begin{equation}
  \rho_{\mathrm{comp}} =
  \operatorname{clip}\left(
    \frac{
      A_{\mathrm{box}} + w_{\mathrm{wall}}A_{\mathrm{wall}}
    }{
      S_{\mathrm{side}}
    },
    0,1
  \right),
\end{equation}
where $w_{\mathrm{wall}}=0.3$ controls the contribution of wall contacts.

The normalized placement height is:
\begin{equation}
  h_{\mathrm{norm}} =
  \operatorname{clip}\left(\frac{z}{R-1},0,1\right).
\end{equation}

Finally, the height-surface fit measures the average height gap between the candidate base plane and the local height map under the footprint, normalized by the item height $d_z$. Here $\eta_{uv}$ denotes the clipped height gap between the placement height $z$ and the local height-map value $H(u,v)$ at footprint cell $(u,v)$. A flatter surface closer to the placement height $z$ yields a value closer to 1.
\begin{align}
  \rho_{\mathrm{fit}}
  &= 1 - \operatorname{clip}\left(
    \frac{1}{d_z |\mathcal{F}(p)|}
    \sum_{(u,v)\in \mathcal{F}(p)} \eta_{uv},
    0,1
  \right), \\
  \eta_{uv}
  &= \operatorname{clip}(z-H(u,v),0,d_z).
\end{align}

\subsection{Hierarchical Policy Network}

The proposed network parameterizes a hierarchical policy, thereby reducing the complexity of learning a single monolithic distribution over item–rotation–position triples. This hierarchy also reflects the natural structure of the packing process: the policy first selects the item best suited to the current packing state and then determines its rotation and placement position.

\paragraph{Spatial Encoder.}
To standardize the experimental comparison environment, we set the container to be cube. We use meters to measure the dimensions of container and items. Any length unit may be used as long as the relative dimensions are preserved. The spatial discretization granularity is controlled by the voxel resolution \(R\). We encode the container state with a voxel-token hybrid spatial encoder, which aggregates local 3D structure with a 3D CNN \citep{3dcnn:3} and models long-range spatial dependencies with a Transformer-based self-attention module (SA1) \citep{tf:4}.

The 3D CNN contains three strided convolutional blocks. Each block $\mathcal{E}_l$ contains two $3 \times 3 \times 3$ convolutions with group normalization and LeakyReLU \citep{lrelu:5}, the first uses stride 2 for downsampling, and the second uses stride 1 for feature refinement. The convolutional hierarchy is
\begin{equation}
  \mathbf{H}_t^1 = \mathcal{E}_1(\mathbf{O}_t), \quad
  \mathbf{H}_t^2 = \mathcal{E}_2(\mathbf{H}_t^1), \quad
  \mathbf{H}_t^3 = \mathcal{E}_3(\mathbf{H}_t^2).
\end{equation}

The channel dimensions are $C=32$, $2C$, and $4C$, respectively, producing a compact feature volume that summarizes local occupancy geometry. We then apply adaptive 3D max pooling to obtain a fixed $N=P \times P \times P$ token grid, $\mathbf{H}_t^{\mathrm{pool}} = \mathrm{Pool}_{\max}(\mathbf{H}_t^3)$, where $d_s=4C$ and $P=4$. The pooled volume is flattened as $\mathbf{S}_t = \mathrm{Flatten}(\mathbf{H}_t^{\mathrm{pool}})^\top$.
To preserve spatial location, each token adds a continuous 3D positional encoding. For token $i$, let
$
\mathbf{p}_i \in [-1,1]^3,
$
be its normalized coordinate on the pooled grid. The initial token is
$
\mathbf{Z}_t^{0}[i] = \mathbf{S}_t[i] + \mathbf{e}_i.
$
Here, $\mathbf{e}_i = \phi_{\mathrm{pos}}(\mathbf{p}_i)$ is produced by a two-layer Multi-Layer Perceptron (MLP) with LeakyReLU. This keeps absolute location information, which is important for space reasoning. Finally, we use self-attention to model global relationships among them.
In our implementation, a single-head, single-layer self-attention (SHA) module keeps the encoder lightweight. The final spatial tokens (ST) are obtained as follows:
\begin{align}
\tilde{\mathbf{Z}}_t
&=
\operatorname{LN}_1\!\left(
\mathbf{Z}_t^{0}
+
\operatorname{SHA}\!\left(\mathbf{Z}_t^{0}\right)
\right), \\
\mathbf{Z}_t
&=
\operatorname{LN}_2\!\left(
\tilde{\mathbf{Z}}_t
+
\operatorname{FFN}\!\left(\tilde{\mathbf{Z}}_t\right)
\right).
\end{align}
Here, LN denotes layer normalization and FFN denotes the feed-forward network. $\mathbf{Z}_t$ is then used by the picker module for item selection.

\paragraph{Picker.}
The picker first encodes the candidate item buffer $B_t$.
Each item $b_{t,i} \in B_t$ is passed through a two-layer MLP
with LeakyReLU activation and projected into the item
embedding space as $\widehat{\mathbf{U}}_t=\phi_{\mathrm{item}}(B_t)$.
To preserve candidate-order information, we add a learnable
order embedding $\mathbf{P}_i$ and apply the buffer-validity mask:
$\mathbf{U}_{t,i}=M_{t,i}^{\mathrm{buf}}
(\widehat{\mathbf{U}}_{t,i}+\mathbf{P}_i)$.

The picker uses item--space cross-attention (CA2) between the candidate
item tokens $\mathbf{U}_t$ and the spatial tokens $\mathbf{Z}_t$. The former
provide queries and the latter provide keys for single-head scaled dot-product
compatibility scoring. Following the pointer mechanism \citep{pointer:6}, we compute
\begin{equation}
\mathbf{Q}_t=\mathbf{U}_t\mathbf{W}^Q, \mathbf{K}_t=\mathbf{Z}_t\mathbf{W}^K, 
\mathbf{G}_t=\mathbf{Q}_t\mathbf{K}_t^\top/\sqrt{d_i},
\end{equation}
where
$g_{t,i,n} \in \mathbf{G}_t$ measures the compatibility between
candidate item $b_{t,i}$ and the $n$-th spatial token
$\mathbf{z}_{t,n}\in \mathbf{Z}_t$. Mean aggregation over the spatial
tokens gives the item logit
$\ell_{t,i}^{\mathrm{item}}
=N^{-1}\sum_{n=1}^{N}g_{t,i,n}$, capturing global item-space compatibility. $\ell_{t,i}^{\mathrm{item}}$ are clipped with tanh, where the range is controlled by hyperparameter $c_{clip}$, and finally normalized by a softmax operation \citep{cclip:7}. 
The resulting
item-selection policy is
\begin{equation}
\pi_{\theta}^{\mathrm{picker}}(b_{t,i}\mid s_t)
=\operatorname{softmax}_{i}
(c_{\mathrm{clip}}\tanh(\ell_{t,i}^{\mathrm{item}})).
\end{equation}

\paragraph{Loader.}
After the picker selects an item $b_{t,i}$, the environment
enumerates feasible rotation--position candidates and filters them using stability check rules. The loader can process up to $J$ candidates. $J$ is adjusted according to the placement generation schemes. If the number of feasible candidates exceeds $J$, orientation-stratified sampling is performed across the $N_{\mathrm{ori}}$ orientations to preserve a balanced orientation distribution after truncation. The resulting
padded candidate set is denoted by
$C_t(b_{t,i})=\{c_{t,j}\}_{j=1}^{J}$. 
$c_{t,j} = (\mathbf{r}_{t,j},\mathbf{p}_{t,j},\mathbf{f}_{t,j}) \in C_t(b_{t,i})$, where $\mathbf{r}_{t,j}=(d_x,d_y,d_z)$ is a rotated-item representation,
$\mathbf{p}_{t,j}=(x,y,z)$ is an anchor coordinate, and
$\mathbf{f}_{t,j}=(\rho_{\mathrm{sup}}, \rho_{\mathrm{comp}}, h_{\mathrm{norm}}, \rho_{\mathrm{fit}})$ is an LGF vector. Then $c_{t,j}$ are embedded and fused into an initial candidate token
\begin{equation}
\mathbf{y}_{t,j}^0=\mathbf{W}^{\mathrm{item}}\phi_{\mathrm{item}}(\mathbf{r}_{t,j})
+\phi_{\mathrm{pos}}(\mathbf{p}_{t,j})
+\phi_{\mathrm{lgf}}(\mathbf{f}_{t,j}).
\end{equation}

The loader uses self-attention (SA2), implemented as two SHA blocks with FFN, to compare feasible placements, masking padded candidates and applying LN after each residual connection \citep{resnet}. For $l\in\{1,2\}$, let $\mathbf{Y}_t^{l-1}=\{\mathbf{y}_{t,j}^{l-1}\}_{j=1}^{J}$,  the candidate representations are updated as 
\begin{align}
\widetilde{\mathbf{Y}}_t^{\,l} &= \operatorname{LN}^{l}(\mathbf{Y}_t^{l-1} + \operatorname{SHA}^{l}(\mathbf{Y}_t^{l-1}; \mathbf{m}_t^{\mathrm{cand}})) \\
\mathbf{Y}_t^{l} &= \operatorname{LN}^{l}(\widetilde{\mathbf{Y}}_t^{\,l} + \operatorname{FFN}^{l}(\widetilde{\mathbf{Y}}_t^{\,l})).
\end{align}

Here, $\mathbf{m}_t^{\mathrm{cand}}$ masks padded candidates.

The default loader omits candidate--space cross-attention (CA1), which would
directly connect placement candidate tokens with spatial tokens
$\mathbf{Z}_t$, following the component analysis in
Section~\ref{subsec:attention-ablation}.

Inspired by the Set Transformer \citep{lee2019set} and Pointer Networks \citep{pointer:6}, the loader scores feasible placements using query and key vectors $\mathbf{q}_{t,j}=\mathbf{W}^{Q}\phi_{\mathrm{item}}(\mathbf{r}_{t,j})$ and $\mathbf{k}_{t,j}=\mathbf{W}^{K}\mathbf{y}_{t,j}^{2}$, yielding the placement logits $\ell_{t,j}^{\mathrm{place}}=\mathbf{q}_{t,j}^{\top}\mathbf{k}_{t,j}/\sqrt{d_p}$. After clipping valid logits and masking padded candidates, the placement policy is
\begin{equation}
\pi_{\theta}^{\mathrm{loader}}(c_{t,j}\mid s_t,b_t)
=\operatorname{softmax}_{j}\!\left(c_{\mathrm{clip}}\tanh(\ell_{t,j}^{\mathrm{place}})\right).
\end{equation}

The selected candidate determines the final item orientation and position.

\subsection{Markov Decision Process Formulation}
We present our problem as Markov Decision Process (MDP).

\paragraph{State.}
The state $s_t\in\mathcal{S}$ at time step $t$ is represented as $s_t=(\mathbf{O}_t,B_t,\mathbf{M}_t^{\mathrm{buf}})$, where $\mathbf{O}_t$ is the voxel occupancy grid, $B_t=\{b_{t,i}\}_{i=1}^{K}$ is the candidate item buffer, and $\mathbf{M}_t^{\mathrm{buf}}$ is the buffer-validity mask. 

\paragraph{Action.}
An action is a two-step hierarchical decision. We denote the joint action as $a_t=(b_{t,i},c_{t,j})\in\mathcal{A}(s_t)$, where the picker first selects a valid item $b_{t,i}\in B_t$ and the loader then selects one feasible placement candidate $c_{t,j}\in\mathcal{C}_t(b_{t,i})=\{c_{t,j}\}_{j=1}^{J}$. The state-dependent joint action space is therefore defined as $\mathcal{A}(s_t)=\{(b_{t,i},c_{t,j})\mid M_{t,i}^{\mathrm{buf}}=1\}$.

\paragraph{Transition.}
The transition $\mathcal{P}(s_{t+1}\mid s_t,a_t)$ is determined by the selected joint action $a_t=(b_{t,i},c_{t,j})$. After feasibility verification, the environment places $b_{t,i}$ according to $c_{t,j}$, updates $\mathbf{O}_t$ to $\mathbf{O}_{t+1}$, removes the selected item from the buffer, adds the next item according to the online arrival process, and refreshes the buffer-validity mask. The episode terminates when no feasible placement $\mathcal{C}_t(b_{t,i}) =\varnothing$ after item selection.

\paragraph{Reward.}
We use a stepwise volume reward.
For each successful placement, the reward is
$
r_t =
\alpha \frac{\operatorname{Vol}(b_{t,i})}{\operatorname{Vol}(\Omega_R)},
$
where $\operatorname{Vol}(\Omega_R)$ is the container volume and $\alpha$ is a scaling coefficient. The immediate reward \(r_t\) measures the utilization gained from the current placement. We use neither terminal rewards nor invalid-action penalties. 
Conditional on successful placement, this reward is identical for all feasible rotation--position candidates of the selected item.
Item selection therefore influences both immediate reward and future packing opportunities, whereas feasible placement choices affect future return by changing the residual geometry without altering the immediate reward.
This reward structure motivates separating item selection from joint rotation--position evaluation.

\subsection{Training Method.}
We optimize hierarchical policy using policy-gradient learning
\citep{williams1992simple}, specifically Proximal Policy Optimization
\citep[PPO;][]{ppo}, an on-policy actor--critic algorithm. The joint policy factorizes as
\begin{equation}
\pi_\theta(a_t\mid s_t)
=
\pi_\theta^{\mathrm{picker}}(b_{t,i}\mid s_t)
\pi_\theta^{\mathrm{loader}}(c_{t,j}\mid s_t,b_{t,i}).
\end{equation}
Thus, the joint action log-probability is the sum of the picker and loader
log-probabilities. The actor $\pi_\theta$ consists of the spatial encoder,
picker, and loader. The critic $V_\psi(s_t)$ uses $\mathbf{Z}_t$ and
$\mathbf{U}_t$ as input and estimates the standard state value of the
hierarchical policy. The same return-based advantage estimate is applied
to the picker and loader log-probabilities, jointly updating item and
placement selection. We collect rollouts from 64 parallel environments
with different random seeds. Each environment is
advanced for $T=5$ steps per rollout, yielding a batch of $5\times64=320$ transitions. We then compute advantages using Generalized Advantage Estimation
\citep[GAE;][]{gae} and perform one PPO update on the collected batch. The actor and critic are trained end-to-end with the clipped policy and value
objectives of PPO. The learning-based methods are trained until there are no significant performance gains.

\begin{table*}
\caption{Performance comparisons. \{util.\} denotes space utilization of the environment (\%); \{std.\} denotes standard deviation of space utilization ($\times 10^{-2}$); \{num.\} denotes the average number of packed items per episode; \{time\} denotes the average decision time per step. SAR1 denotes $K=1$. SAR5 denotes $K=5$. All learning-based methods were set $R=10, N_{\mathrm{ori}}=6$. Bold and underlined entries indicate the best and second-best results within each setting.}
  \label{table1}
\centering
  \begin{tabular*}{\tblwidth}{@{}LRRRRRRRRRRRR@{}}
    \toprule
    & \multicolumn{4}{c}{Dataset 1} & \multicolumn{4}{c}{Dataset 2} & \multicolumn{4}{c}{Dataset 3} \\
    \cmidrule(lr){2-5}\cmidrule(lr){6-9}\cmidrule(lr){10-13}
    Method & util.$\uparrow$ & std.$\downarrow$ & num. & time$\downarrow$ & util.$\uparrow$ & std.$\downarrow$ & num. & time$\downarrow$ & util.$\uparrow$ & std.$\downarrow$ & num. & time$\downarrow$ \\
    \midrule
    \multicolumn{13}{l}{\textit{Heuristic}} \\
    Random & 39.53 & 9.17 & 16.65 & \textless0.1s & 37.30 & 9.33 & 9.84 & \textless0.1s & 34.22 & 10.15 & 7.28 & \textless0.1s \\
    MACS & 63.01 & 9.54 & 24.99 & \textless0.1s & 61.35 & 9.27 & 14.97 & \textless0.1s & 61.96 & 11.09 & 12.41 & \textless0.1s \\
    LSAH & 62.56 & 9.95 & 24.78 & \textless0.1s & 55.82 & 11.49 & 14.27 & \textless0.1s & 56.48 & 9.381 & 11.62 & \textless0.1s \\
    BR & 54.34 & 10.57 & 22.02 & \textless0.1s & 53.65 & 10.21 & 13.39 & \textless0.1s & 54.82 & 10.36 & 11.24 & \textless0.1s \\
    DBL & 68.33 & 9.80 & 26.89 & \textless0.1s & 57.83 & 10.68 & 14.35 & \textless0.1s & 53.55 & 10.82 & 10.76 & \textless0.1s \\
    HM & 64.03 & 8.99 & 25.53 & \textless0.1s & 48.39 & 12.65 & 12.49 & \textless0.1s & 45.08 & 11.96 & 9.14 & \textless0.1s \\
    OnlineBPH & 61.08 & 14.04 & 24.9 & \textless0.1s & 60.27 & 13.04 & 14.92 & \textless0.1s & 59.02 & 11.75 & 11.69 & \textless0.1s \\
    \midrule
    \multicolumn{13}{l}{\textit{Learning-based ($K=1$)}} \\
    Zhao et al. & 62.11 & 11.23 & 24.43 & \textless0.5s & 68.52 & 14.09 & 17.21 & \textless0.5s & 56.33 & 12.19 & 11.79 & \textless0.5s \\
    PCT & 84.75 & 5.35 & 32.44 & \textless0.5s & 79.63 & \textbf{4.61} & 19.35 & \textless0.5s & 77.38 & \textbf{4.66} & 15.10 & \textless0.5s \\
    GOPT & 84.81 & 6.15 & 32.77 & \textless0.5s & 79.25 & 7.18 & 19.29 & \textless0.5s & 76.37 & \underline{6.84} & 14.99 & \textless0.5s \\
    SAR1-EV & 88.78 & \underline{4.79} & 35.29 & \textless0.5s & \textbf{82.97} & 6.96 & \textbf{21.21} & \textless0.5s & \underline{78.19} & 7.35 & \underline{16.55} & \textless0.5s \\
    SAR1-EMS & \underline{88.84} & \textbf{4.65} & \underline{35.32} & \textless0.5s & \underline{81.69} & \underline{6.79} & 20.90 & \textless0.5s & 78.03 & 7.23 & \textbf{16.63} & \textless0.5s \\
    SAR1-FC & \textbf{89.01} & 4.82 & \textbf{35.37} & \textless0.5s & 81.54 & 7.67 & \underline{20.91} & \textless0.5s & \textbf{78.34} & 7.09 & 16.50 & \textless0.5s \\
    \midrule
    \multicolumn{13}{l}{\textit{Learning-based ($K>1$)}} \\
    Yin et al. & 86.53 & 3.51 & 36.01 & \textless0.5s & 88.56 & \textbf{2.49} & 24.29 & \textless0.5s & 87.92 & \textbf{2.86} & \textbf{20.85} & \textless0.5s \\
    SAR5-EV & 92.54 & 4.14 & 37.68 & \textless0.5s & 90.13 & 4.91 & 25.13 & \textless0.5s & \underline{89.30} & 5.35 & 19.90 & \textless0.5s \\
    SAR5-EMS & \underline{92.75} & \textbf{2.91} & \textbf{40.54} & \textless0.5s & \textbf{91.04} & \underline{4.59} & \textbf{25.46} & \textless0.5s & \textbf{90.41} & \underline{4.33} & \underline{20.31} & \textless0.5s \\
    SAR5-FC & \textbf{93.55} & \underline{3.03} & \underline{39.45} & \textless0.5s & \underline{90.96} & 4.75 & \underline{25.16} & \textless0.5s & 89.20 & 4.89 & 19.90 & \textless0.5s \\
    
    \bottomrule
  \end{tabular*}
  
\end{table*}

\section{Experiments}

\subsection{Experimental Setup}

\paragraph{Baselines.}
We conducted comparative experiments on both heuristic and learning-based methods. For the learning-based methods, we compared online and offline approaches. The heuristic methods we used were DBL\citep{dbl:10}, LSAH \citep{lsah:11}, MACS\citep{hu2020tap}, BR\citep{zhao2021online:8}, OnlineBPH\citep{ha2017online}, and HM\citep{wang2019stable}. PCT\citep{pct:1}, GOPT\citep{gopt:9}, and \citet{zhao2021online:8} were online learning methods. \citet{offline} is an offline learning method. Since offline methods allow for the observation of near-complete queue information and support the reordering of the queue, we selected an offline method as a baseline to compare with SAR and evaluate actual performance under a limited observation window. $N_{\mathrm{ori}}$ denotes the number of allowed item orientations.

\paragraph{Datasets.} We designed three independent datasets $\mathcal{D}$ covering both simulated and real-world scenarios. To identify unique item types, we sorted the three dimensions of each item in ascending order and removed duplicate dimension triplets, treating items that differ only by rotation as the same type.
The complete item-type visualization is provided in Figure~\ref{fig:dataset-visualization} in Appendix~\ref{app:dataset-item-types}.
\begingroup
\settowidth{\leftmargini}{\labelitemi}
\addtolength{\leftmargini}{\labelsep}
\begin{itemize}
  \setlength{\itemindent}{0pt}
  \setlength{\listparindent}{0pt}
  \setlength{\parindent}{0pt}
  \item \textbf{Dataset 1} ($\mathcal{D}_1$): Following \citep{zhao2022learning:2}, items are randomly generated with dimensions $(l,w,h)$ satisfying $l,w,h \in \{1,2,3,4,5\}$, include 35 types of items.
  \item \textbf{Dataset 2} ($\mathcal{D}_2$): Data collected from JD Logistics, includes 10 types of items.
  \item \textbf{Dataset 3} ($\mathcal{D}_3$): Data collected from SF Express, includes 11 types of items.
\end{itemize}
\endgroup

\suppressfloats[t]
\begin{figure}
\centering
\begin{tabular}{@{}ccc@{}}
\includegraphics[width=0.30\columnwidth]
  {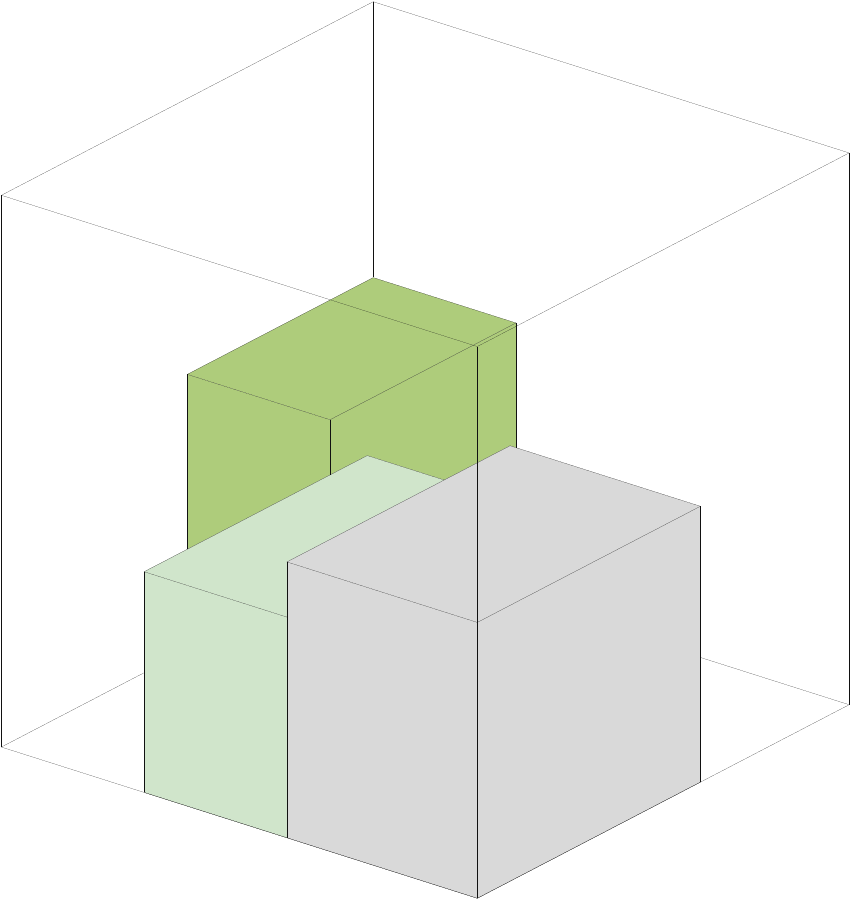}
&
\includegraphics[width=0.30\columnwidth]
  {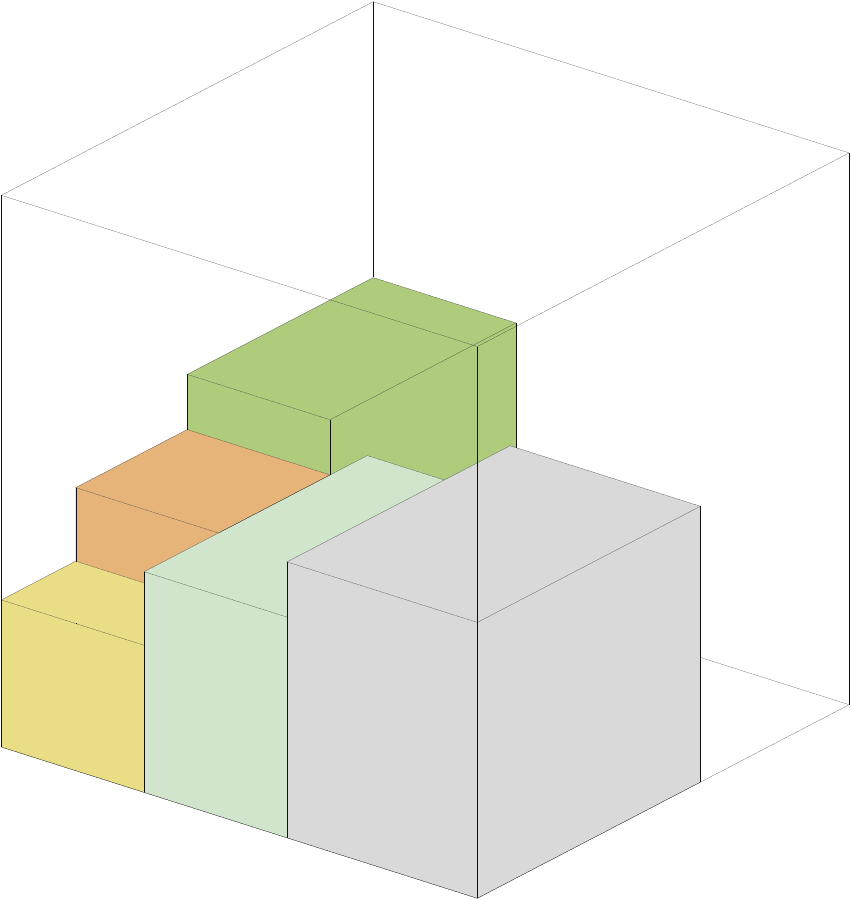}
&
\includegraphics[width=0.30\columnwidth]
  {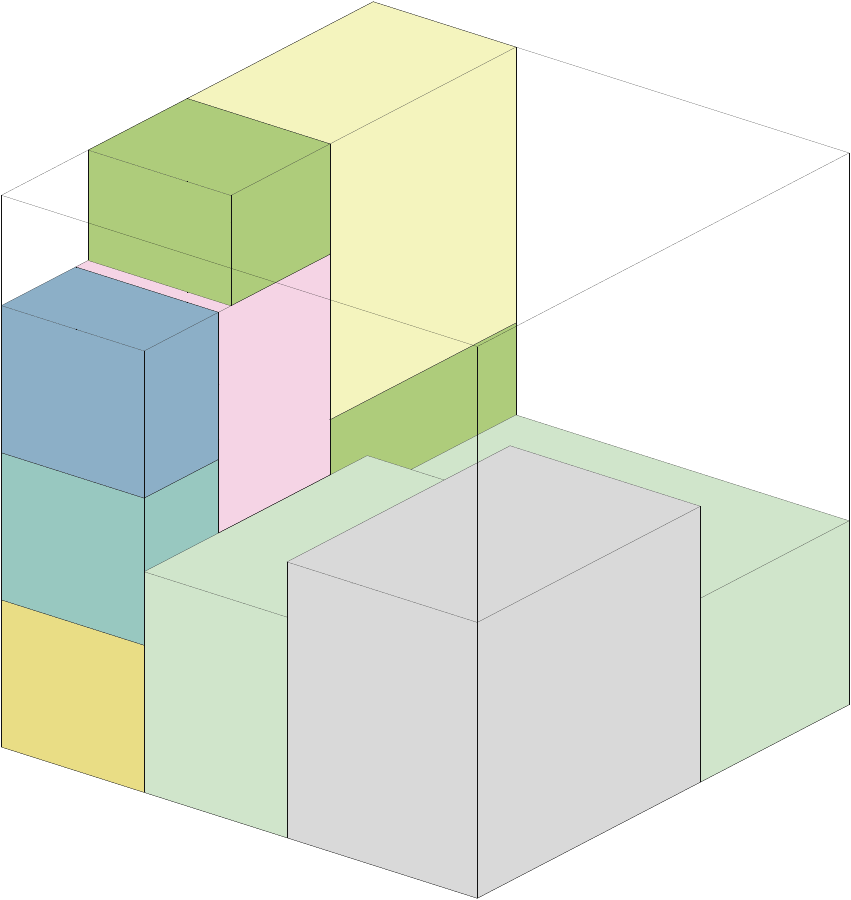}
\\
26.70\%, 3 items
&
31.90\%, 5 items
&
57.13\%, 11 items
\\
\includegraphics[width=0.30\columnwidth]
  {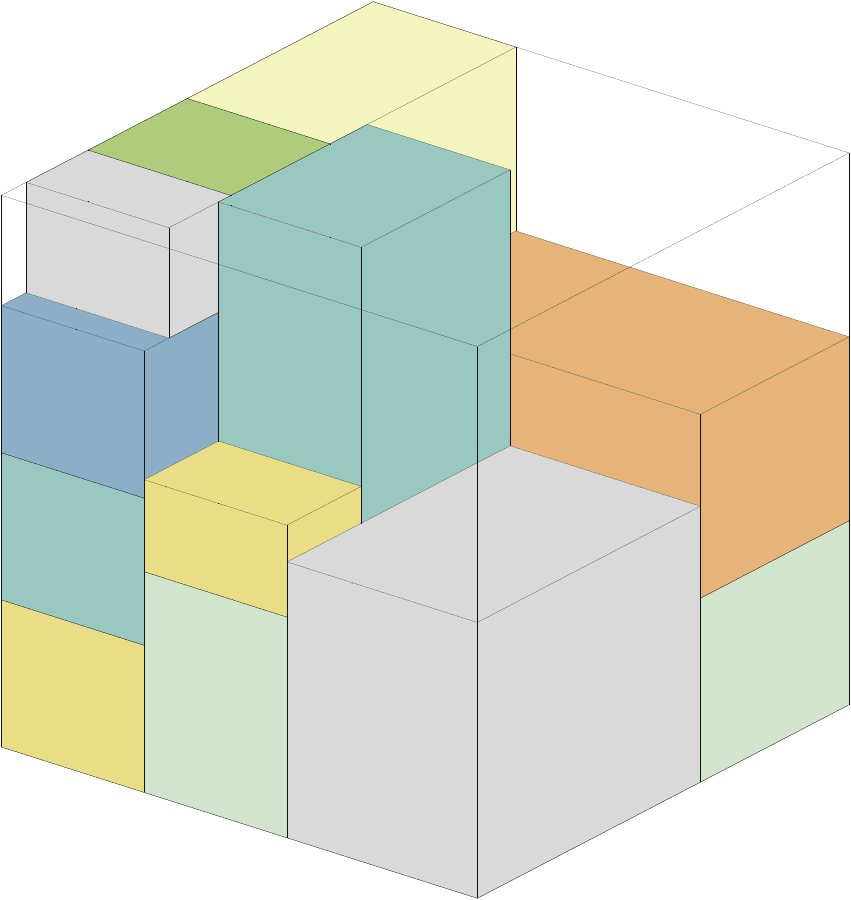}
&
\includegraphics[width=0.30\columnwidth]
  {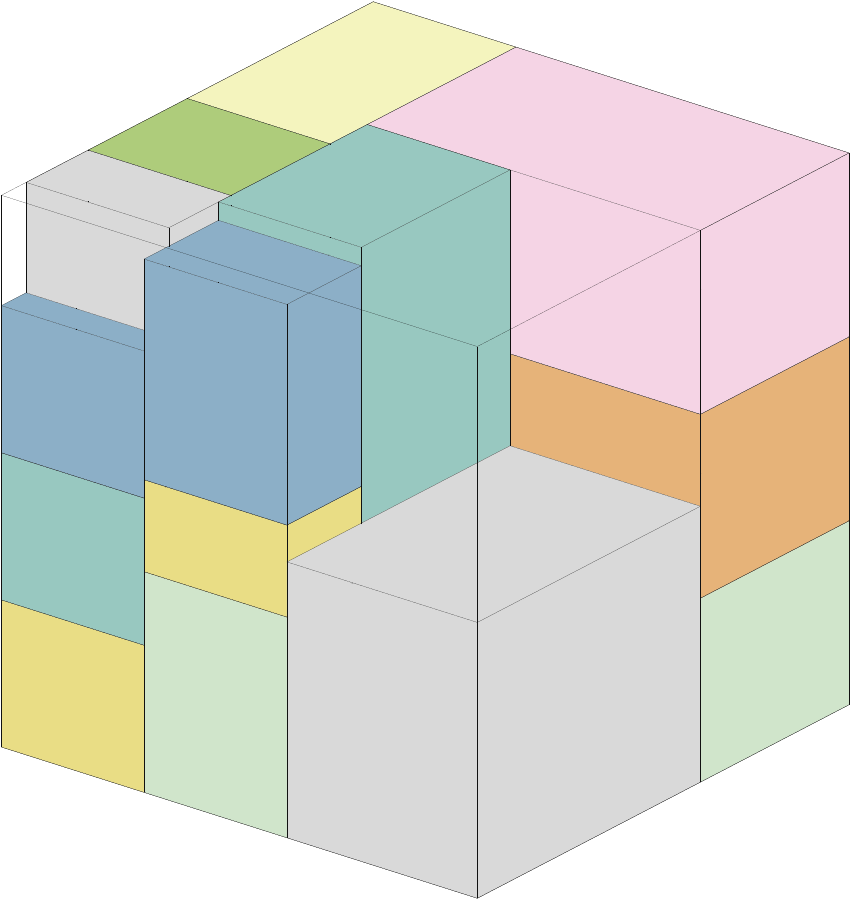}
&
\includegraphics[width=0.30\columnwidth]
  {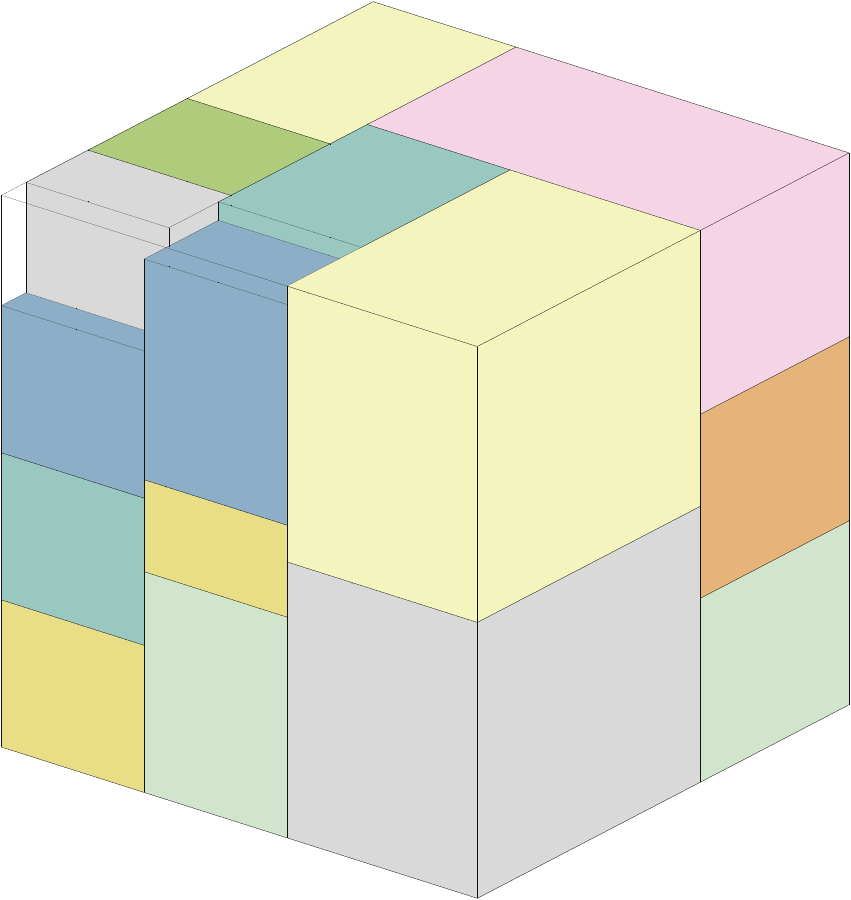}
\\
75.67\%, 15 items
&
87.40\%, 17 items
&
99.40\%, 18 items
\end{tabular}

\caption{Packing progression of one episode on $\mathcal{D}_2$.}
\label{fig:header}
\end{figure}

\begin{figure*}[pos=!t]

\centering
\includegraphics[width=0.98\linewidth]{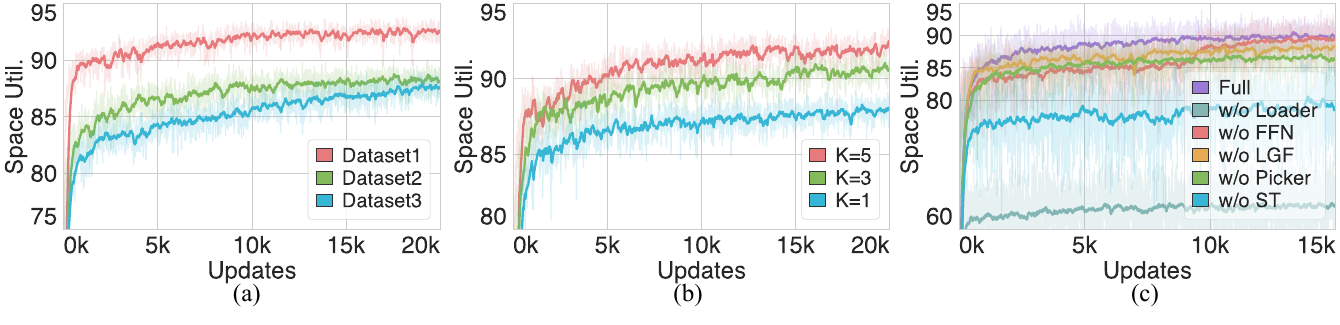}

\caption{Sensitivity Analysis and ablation studies. (a) Performance across the three datasets. (b) Effect of $K$ on $\mathcal{D}_1$. (c) Ablation studies on space utilization during training.}
\label{fig:ablation_ft}
\label{fig:PDT}

\par\vspace{4pt}
\input{sensitivity_ablation_tables.tex}
\end{figure*}

\subsection{Performance Comparison}
We compare SAR with heuristic and learning-based methods over 1,000 episodes with 5 random seeds. The experimental results are shown in Table~\ref{table1}. To match the strict no-buffer setting of PCT and GOPT, SAR1 uses \(K=1\), where the picker has no non-trivial decision, so the comparison mainly evaluates placement selection. Across Datasets 1--3, the best SAR1 variants improve utilization over the strongest \(K=1\) online baseline by 4.20, 3.34, and 0.96 percentage points, respectively.

Figure~\ref{fig:header} shows one packing progression. Increasing the buffer to \(K=5\) further raises utilization to 93.55\%, 91.04\%, and 90.41\% on the three datasets, as shown in Figure~\ref{fig:PDT}(b), reflecting both the larger observation window and learned selection among visible items. We additionally report \citep{offline} as an offline full-information ($K=50$) reference, as it sees a larger item-information horizon than SAR.

Notably, FC constructs feasible candidates by direct coordinate enumeration rather than handcrafted geometric partition rules such as CP, EP, EMS, or EV. The competitive performance of SAR1-FC and SAR5-FC suggests that direct enumeration is a viable candidate-generation option for SAR in the tested settings. The two kinds of generators differ in both candidate coverage and computational cost. Heuristic rules such as EMS and EV shrink the search space to a few promising positions, so fewer candidates have to be generated and ranked at each step, which keeps decisions fast. FC instead enumerates every feasible coordinate and produces a much larger set. Candidate generation and learned ranking thus play complementary roles: generation determines which placement options are available, while ranking evaluates and selects among them, with both components contributing to packing performance. A comparison across CP, EP, EMS, EV, and FC is provided in Section~\ref{subsec:placement-generation-schemes}.

\subsection{Sensitivity Analysis}
Table~\ref{table2} evaluates SAR under different buffer sizes and voxel resolutions, with six item orientations and EV placement generation. On $\mathcal{D}_1$, increasing $K$ from 1 to 5 consistently improves utilization and packed-item count at every tested resolution. For $R=10,30,50$, increasing $K$ from 1 to 3 raises utilization by 3.15, 3.76, and 3.61 percentage points, respectively, whereas increasing $K$ from 3 to 5 adds 0.93, 2.00, and 2.39 percentage points. These gains should be interpreted in the context of 3D-BP: further improvements generally become more difficult as utilization increases, because the remaining free space is limited and often fragmented, making additional items harder to accommodate. Since utilization already exceeds 91\% at $K=3$ for all three tested resolutions, the further improvements at $K=5$ remain meaningful despite their smaller absolute magnitude. A larger buffer exposes more candidate items to the policy and provides more opportunities for reordering, consistent with the training curves in Figure~\ref{fig:PDT}(b). The picker ablation in Table~\ref{table3} further isolates the contribution of learned item reordering.

Voxel resolution has a smaller effect on utilization within the tested range. On $\mathcal{D}_1$, the utilization ranges across $R=10,30,50$ are 0.86, 0.48, and 1.15 percentage points for $K=1,3,5$, respectively, compared with gains of 4.08--6.00 percentage points when increasing $K$ from 1 to 5. Moreover, increasing resolution does not yield monotonic improvements. These results suggest that utilization is more sensitive to buffer size than to voxel resolution under the tested settings.

At $K=5$, the effect of resolution also varies across datasets. Both $\mathcal{D}_1$ and $\mathcal{D}_2$ achieve their highest reported utilization at $R=30$, reaching 93.69\% and 91.80\%, respectively, while $\mathcal{D}_3$ records its highest voxel-based utilization at $R=10$, with 89.30\%. This comparison is affected by the upward-rounding rule used in voxelization: a partially occupied cell is treated as fully occupied, and the reported utilization is calculated from this rounded voxel occupancy rather than the original item volumes. A coarse grid can therefore overestimate occupied volume. Increasing $R$ reduces voxel size and the scale of the rounding error, providing a closer approximation to the actual occupied volume. The higher reported utilization at $R=10$ on $\mathcal{D}_3$ may thus partly reflect this discretization bias. Meanwhile, the packed-item count increases from 19.90 to 20.29 and 21.68 as $R$ increases from 10 to 30 and 50, respectively. These results highlight the need to account for rounding bias when comparing utilization across resolutions; the highest voxel-based utilization alone does not establish the best physical packing performance. No single tested resolution maximizes reported utilization on all three datasets. The training curves in Figure~\ref{fig:PDT}(a) provide a complementary view of the performance differences across datasets.

\subsection{Ablation Studies}

\textbf{Ablation of Components.} Table~\ref{table3} evaluates the contribution of each component. Here, w/o Picker removes learned reordering among buffered items and selects items by a fixed buffer-order rule, while the loader is still used to evaluate placement candidates after an item is chosen. In contrast, w/o Loader keeps item selection but replaces learned ranking over feasible placement candidates with random selection. Removing the loader causes the largest degradation, confirming that candidate-position evaluation is essential beyond feasibility checking. Removing spatial tokens (ST) also substantially reduces utilization, showing that global spatial context benefits item--space matching for reordering decisions. The picker and LGF provide further gains from two different aspects: learned item reordering and fine-grained candidate-position evaluation. Removing FFN has only a minor effect on final performance, but it slows training convergence, as shown in Figure~\ref{fig:PDT}(c). Overall, these results support the complementary roles of item-level reordering and placement-level evaluation in SAR.

\textbf{Attention Architecture Analysis.}
\label{subsec:attention-architecture}
We study how the depth and number of heads in the three attention stages
affect SAR. Table~\ref{tab:attention-architecture} reports results on
$\mathcal{D}_1$ with resolution $R=10$, buffer size $K=5$, all six item
orientations, and FC placement generation. The spatial encoder, picker, and
loader columns give the number of attention heads and layers as H/L.

The reference configuration uses one head and one layer in each attention
stage. Increasing the picker width from one to four heads has almost no effect
on the final policy (93.50\% versus 93.46\% utilization), and widening the
spatial encoder or loader to four heads is not beneficial. In contrast, adding a
second loader layer improves utilization by 0.30 percentage points, reduces
the utilization standard deviation from 3.80 to 3.06, and increases the
average packed-item count by 0.52. This gain is not monotonic: three or four
loader layers return to the reference level, while four or eight heads in the
two-layer loader reduce both utilization and packed-item count. These results
indicate that the loader benefits from a small amount of additional depth for
candidate comparison, but the candidate sets in this task do not require a
wide or very deep attention stack. We therefore retain a two-layer,
single-head loader and lightweight single-head attention in the other stages.

\begin{table}[pos=t,abovecap=0pt]
\caption{Performance under different attention configurations. H/L denotes the
number of attention heads/layers. Util. denotes space utilization (\%),
std. denotes its standard deviation ($\times10^{-2}$), and num. denotes the
average number of packed items per episode.}
\label{tab:attention-architecture}
\setlength{\tabcolsep}{2.5pt}
\begin{tabular*}{\tblwidth}{@{\extracolsep{\fill}}cccccc@{}}
\toprule
Spatial Encoder & Picker & Loader & util.$\uparrow$ & std.$\downarrow$ & num. \\
(H/L) & (H/L) & (H/L) & & & \\
\midrule
1/1 & 1/1 & 1/1 & 93.46 & 3.80 & 38.92 \\
1/1 & 4/1 & 1/1 & 93.50 & 3.52 & 38.95 \\
4/1 & 1/1 & 1/1 & 93.24 & 3.55 & 38.43 \\
1/1 & 1/1 & 4/1 & 93.46 & 3.28 & 38.56 \\
1/2 & 1/1 & 1/1 & 93.05 & 3.66 & 38.98 \\
1/1 & 1/1 & 1/2 & \textbf{93.76} & \textbf{3.06} & \textbf{39.44} \\
1/1 & 1/1 & 1/3 & 93.47 & 3.21 & 38.86 \\
1/1 & 1/1 & 1/4 & 93.25 & 4.28 & 38.28 \\
1/1 & 1/1 & 4/2 & 93.10 & 3.32 & 38.15 \\
1/1 & 1/1 & 8/2 & 93.04 & 3.18 & 38.06 \\
\bottomrule
\end{tabular*}
\end{table}

\begin{figure}[pos=t,abovecap=3pt,belowcap=3pt]
\begingroup
\ExplSyntaxOn
\__reset_tbl:
\skip_zero:N \l_tbl_abovecap_skip
\dim_set:Nn \l_tbl_width_dim { \columnwidth }
\cs_set_eq:cc { @makecaption } { __make_tbl_caption:nn }
\cs_set:cpn { @captype } { table }
\cs_set:Npn \tblwidth { \columnwidth }
\ExplSyntaxOff
\caption{Attention-component ablation. $\checkmark$ and $\times$ indicate
enabled and disabled components, respectively. Each metric is reported as an absolute
result and an adjacent $\Delta$ column showing its change from Full. Util.
denotes space utilization (\%), std. denotes its standard deviation
($\times10^{-2}$), and num. denotes the average packed-item count.}
\label{tab:attention-ablation}
\scriptsize
\setlength{\tabcolsep}{1.5pt}
\begin{tabular*}{\tblwidth}{@{\extracolsep{\fill}}cccccccccc@{}}
\toprule
\multirow{2}{*}{SA1} &
\multirow{2}{*}{SA2} &
\multirow{2}{*}{CA1} &
\multirow{2}{*}{CA2} &
\multicolumn{2}{c}{util.} & \multicolumn{2}{c}{std.} & \multicolumn{2}{c}{num.} \\
\cmidrule(lr){5-6}\cmidrule(lr){7-8}\cmidrule(lr){9-10}
& & & & value & $\Delta$ & value & $\Delta$ & value & $\Delta$ \\
\midrule
$\checkmark$ & $\checkmark$ & $\checkmark$ & $\checkmark$ & 93.39 & (--)    & 4.14 & (--)    & 39.05 & (--) \\
$\times$ & $\checkmark$ & $\checkmark$ & $\checkmark$ & 92.59 & (-0.80) & 4.56 & (+0.42) & 38.24 & (-0.81) \\
$\checkmark$ & $\times$ & $\checkmark$ & $\checkmark$ & 93.24 & (-0.15) & 3.28 & (-0.86) & 38.92 & (-0.13) \\
$\checkmark$ & $\checkmark$ & $\times$ & $\checkmark$ & 93.46 & (+0.07) & 3.80 & (-0.34) & 38.92 & (-0.13) \\
$\checkmark$ & $\times$ & $\times$ & $\checkmark$ & 93.52 & (+0.13) & 3.83 & (-0.31) & 38.69 & (-0.36) \\
$\checkmark$ & $\checkmark$ & $\checkmark$ & $\times$ & 90.07 & (-3.32) & 4.29 & (+0.15) & 34.59 & (-4.46) \\
$\times$ & $\times$ & $\checkmark$ & $\checkmark$ & 93.34 & (-0.05) & 4.08 & (-0.06) & 38.06 & (-0.99) \\
$\checkmark$ & $\checkmark$ & $\times$ & $\times$ & 90.83 & (-2.56) & 3.94 & (-0.20) & 35.57 & (-3.48) \\
$\times$ & $\times$ & $\times$ & $\times$ & 90.23 & (-3.16) & 4.06 & (-0.08) & 34.72 & (-4.33) \\
\bottomrule
\end{tabular*}
\par
\endgroup
\centering
\includegraphics[width=\columnwidth]{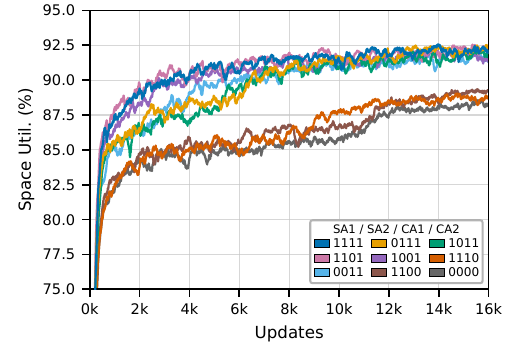}
\caption{Training dynamics for representative attention-component
configurations. Each curve shows the space utilization from one training run
after exponential smoothing with a factor of 0.99.}
\label{fig:attention-component-training}
\end{figure}

\begin{figure}[pos=!t]
\centering
\begin{minipage}[t]{0.48\columnwidth}
\centering
\includegraphics[width=\linewidth]{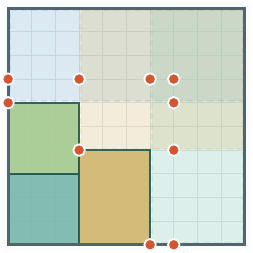}
\small (a) EMS
\end{minipage}
\hfill
\begin{minipage}[t]{0.48\columnwidth}
\centering
\includegraphics[width=\linewidth]{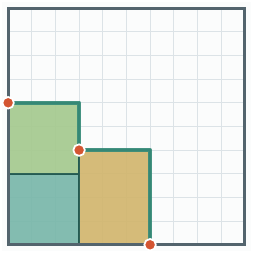}
\small (b) CP
\end{minipage}

\vspace{3pt}

\begin{minipage}[t]{0.48\columnwidth}
\centering
\includegraphics[width=\linewidth]{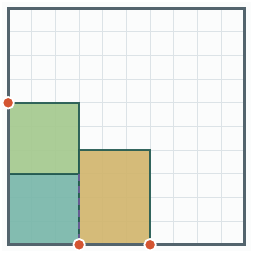}
\small (c) EP
\end{minipage}
\hfill
\begin{minipage}[t]{0.48\columnwidth}
\centering
\includegraphics[width=\linewidth]{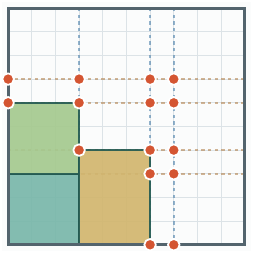}
\small (d) EV
\end{minipage}
\caption{Reference candidate construction on a valid ground-level
packing. (a) EMS aligns the item with the corners of valid empty spaces.
(b) CP extracts positions
where the projected item silhouette transitions from a vertical to a horizontal
boundary segment. (c) EP shows the raw anchors retained by the reference
implementation's sequential 2D projection routine.
(d) EV combines item and container boundary alignments, converted to FLB
coordinates for the candidate footprint, at each relevant boundary height
within the vertical span of $n_t$. Red points denote retained candidate FLB
anchors from the $z=0$ construction plane for an item whose bottom face spans
$3$ grid cells along both the $x$
and $y$ axes (a $3\times3$ footprint). These anchors are shown before height
lifting and physics checks; the EP anchor at $(3,0)$ is lifted above the
yellow item.}
\label{fig:reference-candidate-generators}
\end{figure}

\textbf{Ablation of Attention Components.}
\label{subsec:attention-ablation}
We further isolate the contribution of each attention component under
$R=10$, $K=5$, six item orientations, and FC placement generation.
Table~\ref{tab:attention-ablation} ablates spatial encoder self-attention (SA1),
loader self-attention (SA2), loader candidate--space cross-attention (CA1), and
picker item--space cross-attention (CA2). CA1 performs cross-attention between
candidate placement tokens and spatial tokens, allowing placement evaluation
to incorporate global container context. A checkmark ($\checkmark$) indicates that the
component is enabled, whereas a cross ($\times$) indicates that it is disabled.
The adjacent $\Delta$ column gives the change relative to the Full model.

\begin{figure}[pos=!t]
\centering
\begin{minipage}[t]{0.48\columnwidth}
\centering
\includegraphics[width=\linewidth]{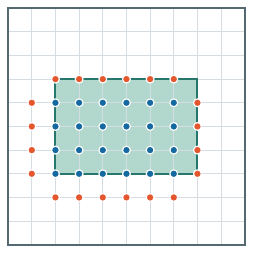}
\small (a) V1
\end{minipage}
\hfill
\begin{minipage}[t]{0.48\columnwidth}
\centering
\includegraphics[width=\linewidth]{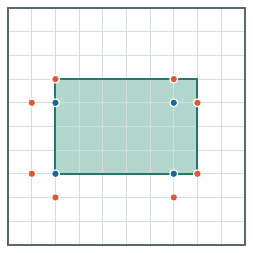}
\small (b) V2
\end{minipage}
\caption{Top-view illustration of the incremental candidate-generation
variants. Blue points denote top-face candidates, and orange points denote
bottom-perimeter candidates. Points lie on integer grid intersections and
show the $(x,y)$ coordinates of voxel lower corners used as candidate FLB
anchors for subsequent items. V1 densely samples the top face and perimeter
edges, whereas V2 retains only their corner and endpoint representatives.}
\label{fig:v1-v2-candidates}

\par\vspace{6pt}
\begingroup
\ExplSyntaxOn
\__reset_tbl:
\skip_zero:N \l_tbl_abovecap_skip
\dim_set:Nn \l_tbl_width_dim { \columnwidth }
\cs_set_eq:cc { @makecaption } { __make_tbl_caption:nn }
\cs_set:cpn { @captype } { table }
\cs_set:Npn \tblwidth { \columnwidth }
\ExplSyntaxOff
\sffamily\small
\caption{Performance of SAR with different placement generation
schemes. Util. denotes space utilization (\%), std. denotes its standard
deviation ($\times10^{-2}$), num. denotes the average number of packed
items per episode, and time denotes the average decision time per step (ms).}
\label{tab:placement-generation-schemes}
\setlength{\tabcolsep}{3pt}
\begin{tabular*}{\tblwidth}{@{\extracolsep{\fill}}ccccc@{}}
\toprule
Scheme & util.$\uparrow$ & std.$\downarrow$ & num. & time \\
\midrule
EMS & 92.75 & \textbf{2.91} & \textbf{40.54} & 126 \\
CP  & 89.42 & 4.16 & 36.34 & 90 \\
EV  & 93.17 & 3.42 & 38.78 & 279 \\
FC  & \textbf{93.46} & 3.00 & 39.45 & 300 \\
EP  & 89.06 & 5.12 & 36.84 & 116 \\
V1  & 93.28 & 3.68 & 38.75 & 280 \\
V2  & 92.42 & 4.31 & 38.20 & 185 \\
\bottomrule
\end{tabular*}
\par
\endgroup
\end{figure}

The component ablation separates the global and local roles of the attention
blocks. Removing picker item--space cross-attention (CA2) causes the largest single
drop, decreasing utilization by 3.32 percentage points and the packed-item
count by 4.46. This is consistent with the picker implementation: without
CA2, item scores cannot directly condition on the current spatial tokens.
Removing spatial encoder self-attention (SA1) also has a clear cost (0.80 percentage
points and 0.81 items), although the loss is smaller than for CA2. By
comparison, removing loader self-attention (SA2) or loader cross-attention (CA1) changes
utilization by at most 0.15 percentage points in this matrix; disabling both
even gives a small improvement over the all-enabled reference. Thus the
additional global context supplied by CA1 does not improve utilization in
this setting, whereas global item--space interaction through CA2 is central
to the learned policy.

Table~\ref{tab:attention-ablation} and
Figure~\ref{fig:attention-component-training} show that the variant without
CA1 offers the most favorable overall balance of packing performance and
convergence behavior among the tested configurations. It maintains high
utilization and packed-item count, reduces utilization variability relative
to the all-enabled model, and shows rapid initial learning. Some alternatives
improve an individual final metric, but yield fewer packed items or slower
initial learning. Considering packing quality, variability, and convergence
together, we therefore adopt the variant without CA1 as the default network
configuration for our experiments.

\subsection{Placement Generation Schemes}
\label{subsec:placement-generation-schemes}

We further evaluate whether SAR depends on a particular heuristic for
generating placement candidates. Table~\ref{tab:placement-generation-schemes}
reports results on $\mathcal{D}_1$ using a $10\times10\times10$ container,
a visible-item buffer of $K=5$, and all six item orientations. We compare
Empty Maximal Space (EMS), Corner Point (CP), Event Point (EV), Full
Coordinate (FC), Extreme Point (EP), and two incremental candidate-generation
variants (V1 and V2). All other model and environment settings follow the
defaults used elsewhere in this section.

\begin{figure}[pos=!t]
\centering
\includegraphics[width=\columnwidth]{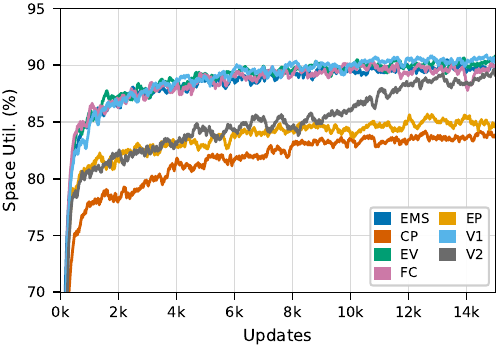}
\caption{Training dynamics under different placement-generation
schemes. Each curve shows the space utilization from one training run after
exponential smoothing with a factor of 0.99. Final policy evaluation results
are reported in Table~\ref{tab:placement-generation-schemes}.}
\label{fig:placement-training}
\end{figure}

\textbf{Classical geometric schemes.}
EMS maintains a set of maximal axis-aligned empty cuboids, each defined by its
front-left-bottom (FLB) corner and dimensions. When item $n_t$ is placed, every
EMS that intersects $n_t$ is split into residual EMSs; any EMS fully
contained within another is then removed. For every EMS large enough to hold the
current item orientation, candidates are formed by aligning the item footprint
with the left-up, right-up, left-bottom, and right-bottom corners of the EMS
vertical bottom face, converted to FLB placement coordinates, as shown in
Figure~\ref{fig:reference-candidate-generators}(a).

CP identifies positions where the silhouette of packed items, projected onto a
2D axis plane, transitions from a vertical to a horizontal boundary segment,
as shown in Figure~\ref{fig:reference-candidate-generators}(b).
When $n_t$ is placed, 3D candidate positions are introduced from the set
$\{(p_n^x{+}s_n^x,\,p_n^y,\,p_n^z),\,\allowbreak(p_n^x,\,p_n^y{+}s_n^y,\,p_n^z),\,\allowbreak(p_n^x,\,p_n^y,\,p_n^z{+}s_n^z)\}$
on whichever of the $xoy$, $yoz$, and $xoz$ planes has its boundary envelope
changed by $n_t$. Positions that no longer satisfy the envelope condition are
removed.

EP follows the sequential 2D projection routine in the reference implementation
of \citet{pct:1}. At ground level and each distinct item-top height, projected
items are sorted by increasing lower $y$ coordinate, with increasing upper $x$
coordinate breaking ties. Each item's upper-left and lower-right boundary
points are projected toward the negative $x$ and $y$ directions, respectively,
onto eligible faces of previously processed items or container walls.
Previously collected anchors covered by the current item are removed before
its new projections are added. The routine retains newly appearing anchors
relative to the preceding height plane. These raw anchors can overlap another
item's footprint in top view, as at $(3,0)$ in
Figure~\ref{fig:reference-candidate-generators}(c); height lifting and physical
checks subsequently determine their placement heights and feasibility.

EV combines item and container boundary alignments to generate candidate
positions \citep{pct:1}. When $n_t$ is placed, the existing
candidate set is updated by scanning distinct boundary heights $p^z$, given by
the start/end $z$ coordinates of packed items and satisfying
$p_n^z \le p^z \le p_n^z+s_n^z$. In each corresponding $z$-plane, the
start/end $x$ and $y$ coordinates of $n_t$ are combined with existing item
boundaries and container boundaries; the initial empty container is treated as
a special packed item. For each orientation of the item to be placed next,
boundary alignments are converted to FLB coordinates using its horizontal
dimensions $(s^x,s^y)$. An edge aligned to a boundary coordinate $b_x$ or $b_y$
gives an FLB coordinate of $b_x$ or $b_x-s^x$ along $x$, and $b_y$ or
$b_y-s^y$ along $y$, depending on which edge is aligned. Cartesian combinations
of the resulting $x$ and $y$ anchors form proposals for that plane.
The reference generator retains proposals whose complete footprints lie within
an EMS at the same height, as shown in
Figure~\ref{fig:reference-candidate-generators}(d).

\begin{figure}[pos=!t]
\centering
\includegraphics[width=0.8\columnwidth]{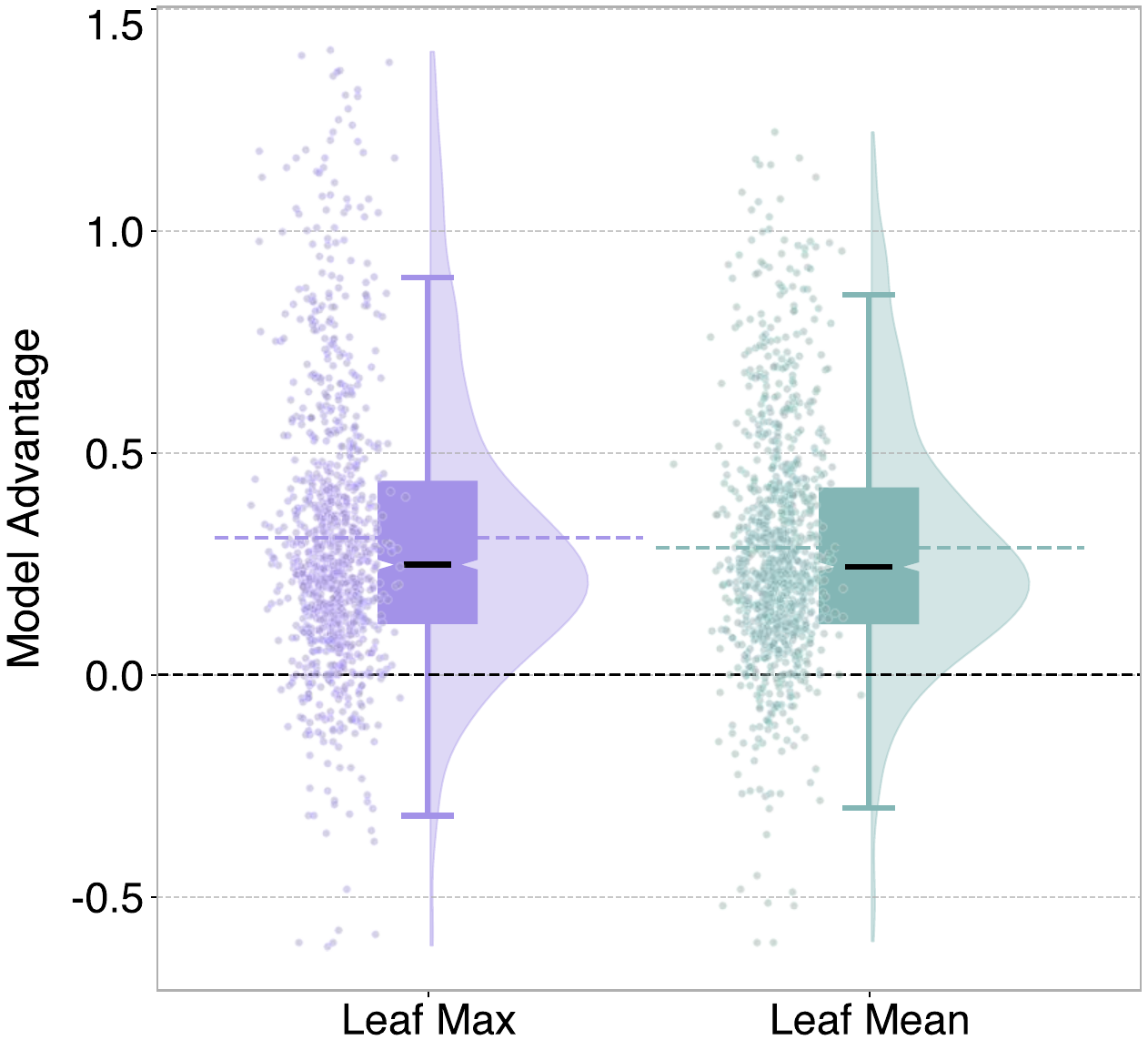}
\caption{Episode-averaged model advantage (percentage points) under Leaf-Max
and Leaf-Mean. Positive and negative values favor the selected and unselected
items, respectively; zero denotes equal outcomes.}
\label{fig:local-advantage}
\end{figure}

Figure~\ref{fig:reference-candidate-generators} illustrates the four
constructions on a common top-view state, with the EMSs and event-boundary
records updated after each placed item. In the implementation, every
two-dimensional anchor is lifted to the maximum occupied height below the
candidate footprint. Boundary, collision, and stability checks then produce the
feasible placement set passed to the loading policy.

\textbf{Incremental variants.}
To assess whether SAR depends on carefully designed geometric
candidate-generation rules, we compare EMS, CP, EP, and EV with two simple
incremental variants based on local sampling around newly placed items.
Figure~\ref{fig:v1-v2-candidates} illustrates their candidate anchors.
V1 and V2 update a persistent candidate set after each placement. For a placed
item with lower corner $(x,y,z)$ and voxel dimensions $(d_x,d_y,d_z)$, define
$x_0=x$, $x_1=x+d_x-1$, $y_0=y$, $y_1=y+d_y-1$, and
$z_{\mathrm{top}}=z+d_z$. V1 adds every empty voxel on the top face together
with the positions adjacent to the four bottom perimeter edges. V2 sparsifies
this set to four top corners and eight bottom-perimeter points:
\begin{align}
\mathcal{C}^{\mathrm{V2}}_{\mathrm{top}}
&=\{(x_i,y_j,z_{\mathrm{top}})\mid i,j\in\{0,1\}\},\notag\\
\mathcal{C}^{\mathrm{V2}}_{\mathrm{bottom}}
&=\begin{aligned}[t]
&\{(x_0,y_1{+}1,z),(x_0{-}1,y_1,z),\\
&\quad (x_0{-}1,y_0,z),(x_0,y_0{-}1,z),\\
&\quad (x_1,y_0{-}1,z),(x_1{+}1,y_0,z),\\
&\quad (x_1{+}1,y_1,z),(x_1,y_1{+}1,z)\}.
\end{aligned}
\label{eq:v2-candidates}
\end{align}
The accumulated candidates are deduplicated and filtered by boundary,
collision, and stability checks before placement selection.

The competitive performance of V1 and V2 in
Table~\ref{tab:placement-generation-schemes} further supports flexible
customization of candidate-generation rules to application requirements while
keeping SAR's policy architecture unchanged.

\subsection{Behavioral Analysis}
To examine whether learned item selection improves packing outcomes in
crowded states, we conduct a counterfactual tree analysis on $\mathcal{D}_1$
($K{=}2$, $N_{\mathrm{ori}}{=}6$, FC) over 1{,}000 episodes.
Before utilization reaches 85\%, packing follows only the model-selected
branch; thereafter, both the selected and unselected item branches are
expanded recursively. Their sibling subtrees start from the same packing
state, enabling a controlled comparison of alternative item choices.
Leaf-Max and Leaf-Mean denote, respectively, the maximum and mean utilization
over the leaves of each explored subtree.

For each comparison node, we define the \emph{model advantage} as
$\Delta_q = U_q^{\mathrm{sel}} - U_q^{\mathrm{unsel}}$,
where $q\in\{\mathrm{max},\mathrm{mean}\}$ specifies the leaf statistic,
and $U_q^{\mathrm{sel}}$ and $U_q^{\mathrm{unsel}}$ are the corresponding
utilization values for the selected and unselected branches.
The difference is measured in percentage points: a positive value indicates
higher utilization for the model-selected branch, zero indicates equal
outcomes, and a negative value indicates higher utilization for the
unselected branch. Figure~\ref{fig:local-advantage} shows the distributions
of these differences averaged over comparison nodes within each episode.
Thus, the sign of a plotted value describes an episode-level average rather
than every individual decision.

The win/tie/loss ratios are 88.66/1.05/10.29~(\%) for Leaf-Max
and 90.02/0.84/9.14~(\%) for Leaf-Mean.
Leaf-Max characterizes the best packing outcome found within an explored
subtree, and Leaf-Mean characterizes the average outcome across its explored
leaves. The high win rates under both statistics suggest that learned item
selection favors branches with better best-case outcomes and higher average
leaf utilization. Together, these results support the effectiveness of learned
item selection in improving packing outcomes under the tested crowded conditions.

\subsection{Adaptation Ability}
\label{subsec:adaptation-ability}

We evaluate cross-dataset transferability through zero-shot evaluation and
sequential fine-tuning. Model A is trained on Dataset 1 and directly evaluated
on Datasets 2 and 3 without parameter updates, yielding the zero-shot results.
We then fine-tune Model A on Dataset 2 to obtain Model B and further fine-tune
Model B on Dataset 3 to obtain Model C. The fine-tuned results are obtained by
evaluating Model C on all three datasets. Model A's performance on Dataset 1
is also reported as a source-domain reference.

\edef\sarRestoreTopFloatLimit{\noexpand\setcounter{topnumber}{\arabic{topnumber}}}
\setcounter{topnumber}{1}
\AddToHookNext{cmd/onecolumn/after}{\sarRestoreTopFloatLimit}
\begin{table}[pos=t,abovecap=0pt]
\caption{Generalization performance under zero-shot and fine-tuned settings.
Util. denotes space utilization (\%), std. denotes its standard deviation
($\times10^{-2}$), and num. denotes the average packed-item count. Underlined
values correspond to source-domain evaluation of Model A.}
\label{tab:adaptation-ability}
\scriptsize
\setlength{\tabcolsep}{1.5pt}
\begin{tabular*}{\tblwidth}{@{\extracolsep{\fill}}lccccccccc@{}}
\toprule
& \multicolumn{3}{c}{Dataset 1} & \multicolumn{3}{c}{Dataset 2} & \multicolumn{3}{c}{Dataset 3} \\
\cmidrule(lr){2-4}\cmidrule(lr){5-7}\cmidrule(lr){8-10}
Method & util.$\uparrow$ & std.$\downarrow$ & num. & util.$\uparrow$ & std.$\downarrow$ & num. & util.$\uparrow$ & std.$\downarrow$ & num. \\
\midrule
zero-shot & \underline{93.69} & \underline{3.88} & \underline{37.99} & 76.44 & 8.00 & 22.24 & 72.34 & 8.47 & 17.56 \\
fine-tuned & 87.42 & 5.72 & 37.24 & 89.22 & 4.65 & 25.04 & 86.69 & 5.57 & 21.53 \\
\bottomrule
\end{tabular*}
\end{table}

\suppressfloats[t]
\begin{figure}[pos=t]
\centering
\includegraphics[width=\linewidth]{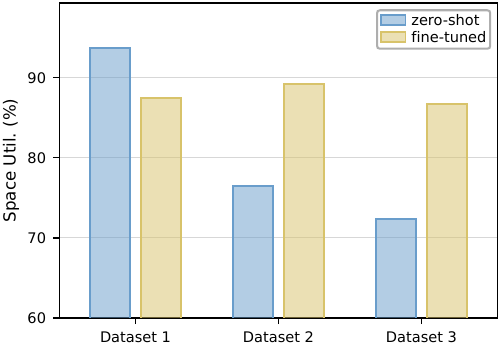}
\caption{Cross-dataset space utilization under zero-shot and fine-tuned
settings.}
\label{fig:adaptation-utilization}
\end{figure}

Table~\ref{tab:adaptation-ability} and Figure~\ref{fig:adaptation-utilization}
show that the final fine-tuned model improves utilization on Dataset 2 from
76.44\% to 89.22\% and on Dataset 3 from 72.34\% to 86.69\%, relative to
zero-shot evaluation. These increases amount to 12.78 and 14.35 percentage
points, respectively. The utilization standard deviation decreases from
8.00 to 4.65 on Dataset 2 and from 8.47 to 5.57 on Dataset 3, while the average
packed-item count increases from 22.24 to 25.04 and from 17.56 to 21.53,
respectively. Thus, the utilization gains are accompanied by more packed items
and lower utilization variability on both target datasets.

On Dataset 1, utilization decreases from 93.69\% for Model A to 87.42\% for
Model C, a loss of 6.27 percentage points, and the average packed-item count
decreases from 37.99 to 37.24. These results support SAR's adaptability under
the tested Dataset 1 $\rightarrow$ Dataset 2 $\rightarrow$ Dataset 3
fine-tuning sequence, with a trade-off between target-domain improvement and
source-domain performance retention. The findings characterize the final
Model C obtained after this sequence.

\section{Conclusion}
This paper presented SAR, a learning-based framework for buffered online 3D-BP.
This decision decomposition reflects the distinct roles of item selection and rotation--position selection: conditional on successful placement, item selection determines the immediate volume gain and also influences future returns, while rotation and position shape the remaining free-space configuration, thereby affecting subsequent packing opportunities and future returns.
The picker selects buffered items using voxel-based spatial tokens, while the loader performs fine-grained evaluation of feasible rotation--position candidates using LGF and inter-candidate comparison.
Both stages are jointly optimized under a shared long-term packing return.
SAR improves utilization over the evaluated online baselines without buffering, with further gains when a buffer is available.
Ablations support the complementary contributions of voxel-based item selection and LGF-based placement evaluation, while experiments demonstrate compatibility with multiple candidate generators.
Counterfactual analysis further supports the benefits of learned item selection in crowded packing states.

Several limitations remain.
SAR assumes cuboid items and standardized cubic containers, with evaluation limited to simulation.
Distribution shifts reduce zero-shot performance, while fine-tuning involves a trade-off between adaptation and retention.
Voxel rounding may also bias reported utilization measurements.
Future work should extend geometric applicability, improve adaptation and retention, validate physical deployment, and refine volume evaluation.
We hope that our work could provide a new perspective from which to formulate and examine 3D-BP and related problems.

\onecolumn
\appendix
\setcounter{figure}{0}
\renewcommand{\thefigure}{\thesection.\arabic{figure}}
\renewcommand{\theHfigure}{appendix.\thesection.\arabic{figure}}
\begingroup
\makeatletter
\renewcommand{\@seccntformat}[1]{\appendixname~\csname the#1\endcsname.\hskip 0.5em}
\makeatother
\section{Dataset Item Types}
\label{app:dataset-item-types}
\endgroup

\def\dslabel#1-#2-#3\relax{(#1,#2,#3)}
\newlength{\dscolumnwidth}

\begin{figure*}[pos=!ht]
\centering

\begingroup
\centering
\setlength{\tabcolsep}{0pt}
\begin{tabular*}{\textwidth}{@{\extracolsep{\fill}}ccccccc@{}}
  \begin{minipage}[t][0.85in][t]{0.13\textwidth}
    \centering
    \includegraphics[width=\linewidth,height=0.69in,keepaspectratio]{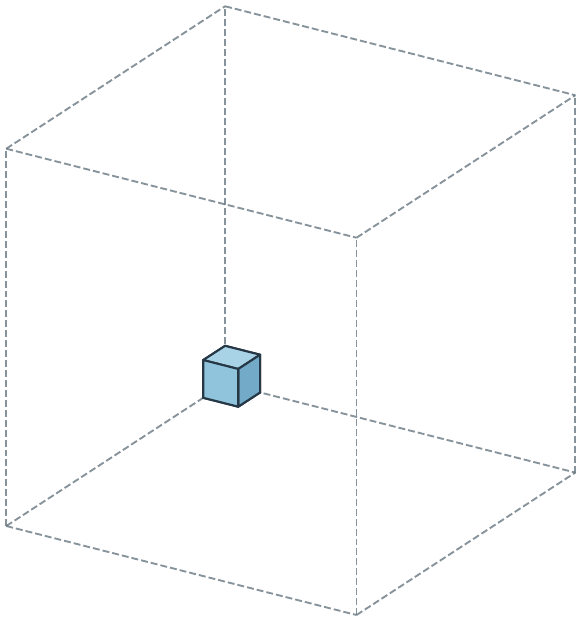}%
    \par\vspace{2pt}
    {\footnotesize \dslabel1-1-1\relax}
  \end{minipage}%
 &
  \begin{minipage}[t][0.85in][t]{0.13\textwidth}
    \centering
    \includegraphics[width=\linewidth,height=0.69in,keepaspectratio]{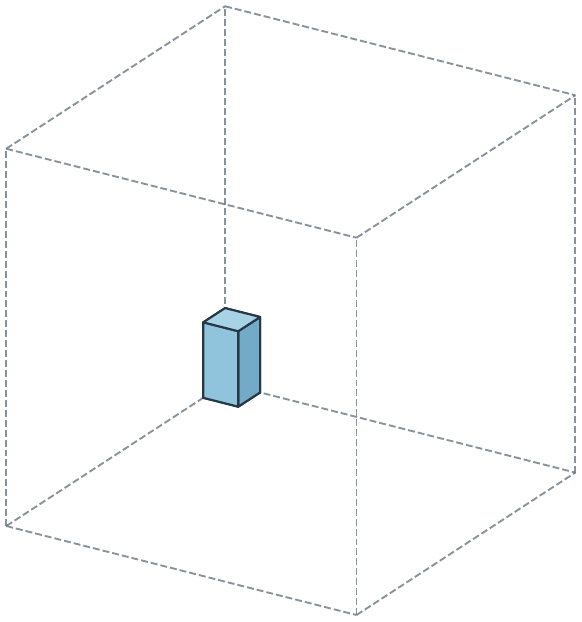}%
    \par\vspace{2pt}
    {\footnotesize \dslabel1-1-2\relax}
  \end{minipage}%
 &
  \begin{minipage}[t][0.85in][t]{0.13\textwidth}
    \centering
    \includegraphics[width=\linewidth,height=0.69in,keepaspectratio]{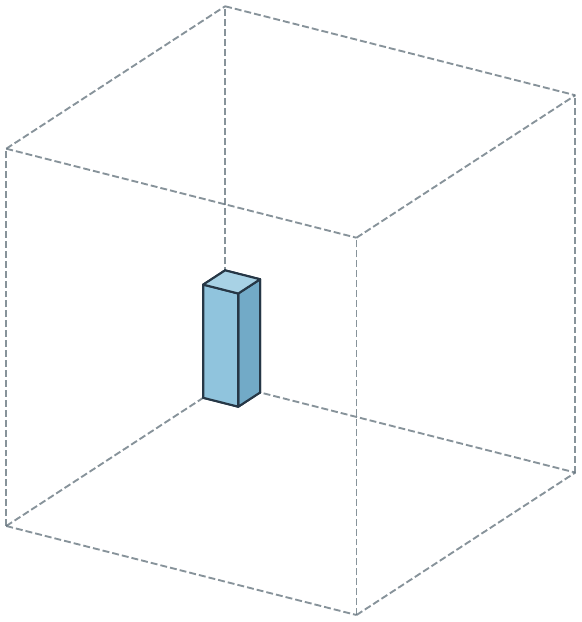}%
    \par\vspace{2pt}
    {\footnotesize \dslabel1-1-3\relax}
  \end{minipage}%
 &
  \begin{minipage}[t][0.85in][t]{0.13\textwidth}
    \centering
    \includegraphics[width=\linewidth,height=0.69in,keepaspectratio]{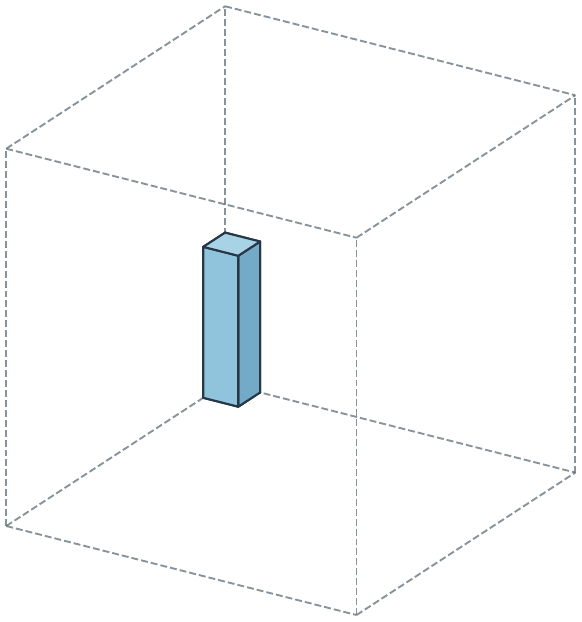}%
    \par\vspace{2pt}
    {\footnotesize \dslabel1-1-4\relax}
  \end{minipage}%
 &
  \begin{minipage}[t][0.85in][t]{0.13\textwidth}
    \centering
    \includegraphics[width=\linewidth,height=0.69in,keepaspectratio]{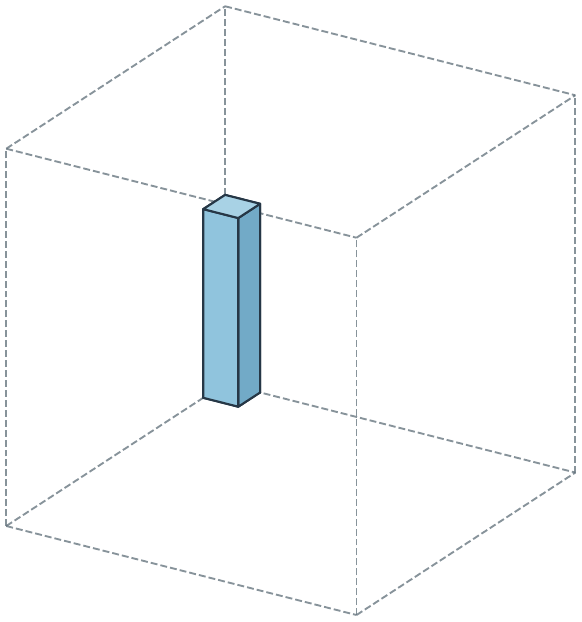}%
    \par\vspace{2pt}
    {\footnotesize \dslabel1-1-5\relax}
  \end{minipage}%
 &
  \begin{minipage}[t][0.85in][t]{0.13\textwidth}
    \centering
    \includegraphics[width=\linewidth,height=0.69in,keepaspectratio]{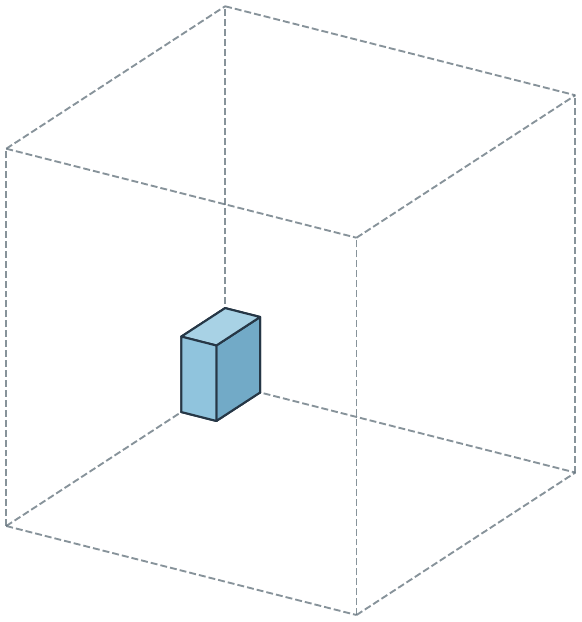}%
    \par\vspace{2pt}
    {\footnotesize \dslabel1-2-2\relax}
  \end{minipage}%
 &
  \begin{minipage}[t][0.85in][t]{0.13\textwidth}
    \centering
    \includegraphics[width=\linewidth,height=0.69in,keepaspectratio]{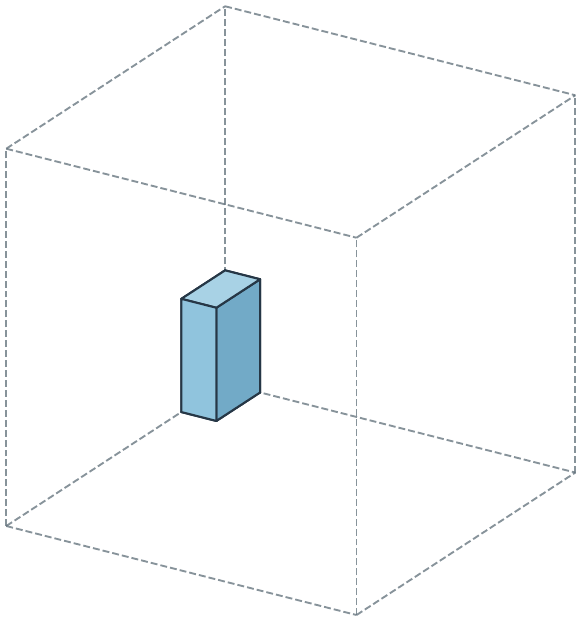}%
    \par\vspace{2pt}
    {\footnotesize \dslabel1-2-3\relax}
  \end{minipage}%
 \\
  \begin{minipage}[t][0.85in][t]{0.13\textwidth}
    \centering
    \includegraphics[width=\linewidth,height=0.69in,keepaspectratio]{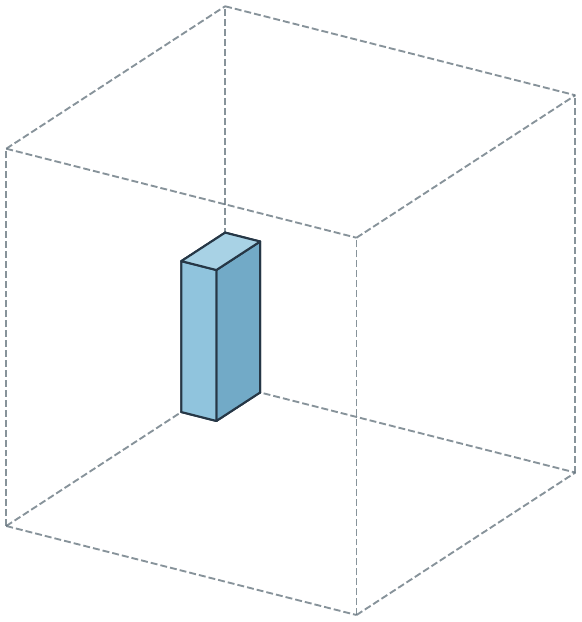}%
    \par\vspace{2pt}
    {\footnotesize \dslabel1-2-4\relax}
  \end{minipage}%
 &
  \begin{minipage}[t][0.85in][t]{0.13\textwidth}
    \centering
    \includegraphics[width=\linewidth,height=0.69in,keepaspectratio]{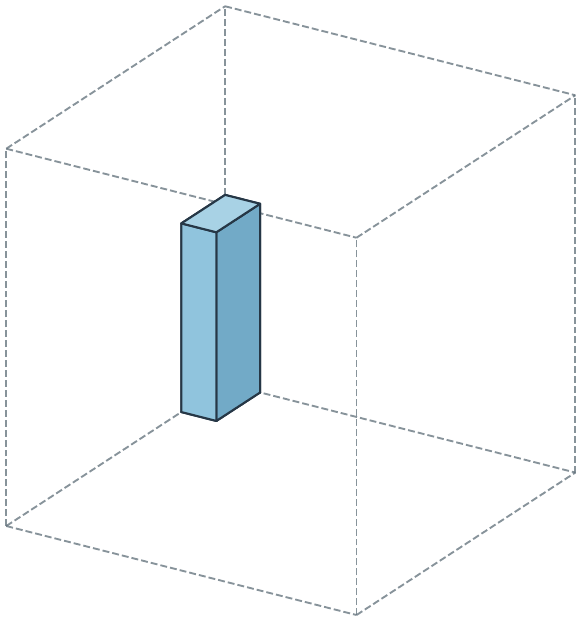}%
    \par\vspace{2pt}
    {\footnotesize \dslabel1-2-5\relax}
  \end{minipage}%
 &
  \begin{minipage}[t][0.85in][t]{0.13\textwidth}
    \centering
    \includegraphics[width=\linewidth,height=0.69in,keepaspectratio]{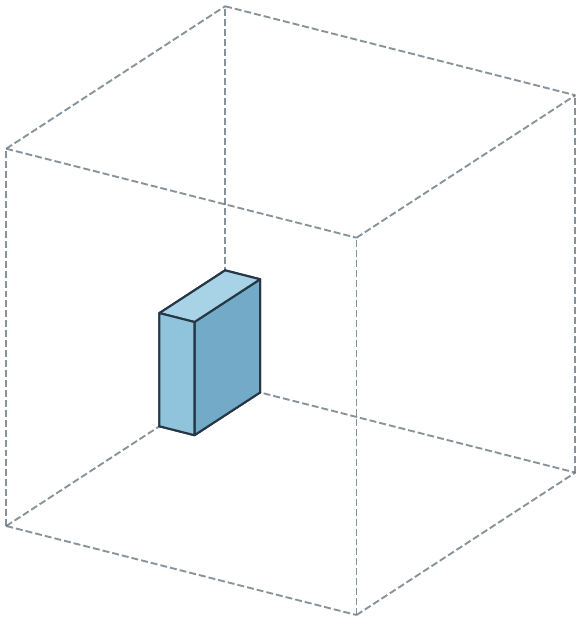}%
    \par\vspace{2pt}
    {\footnotesize \dslabel1-3-3\relax}
  \end{minipage}%
 &
  \begin{minipage}[t][0.85in][t]{0.13\textwidth}
    \centering
    \includegraphics[width=\linewidth,height=0.69in,keepaspectratio]{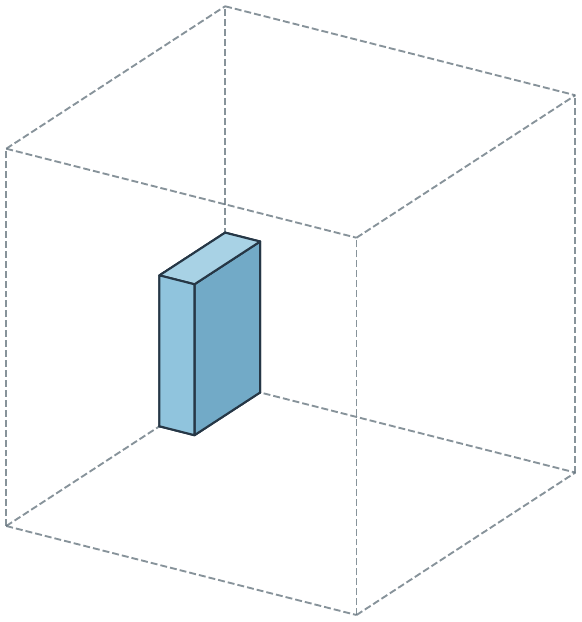}%
    \par\vspace{2pt}
    {\footnotesize \dslabel1-3-4\relax}
  \end{minipage}%
 &
  \begin{minipage}[t][0.85in][t]{0.13\textwidth}
    \centering
    \includegraphics[width=\linewidth,height=0.69in,keepaspectratio]{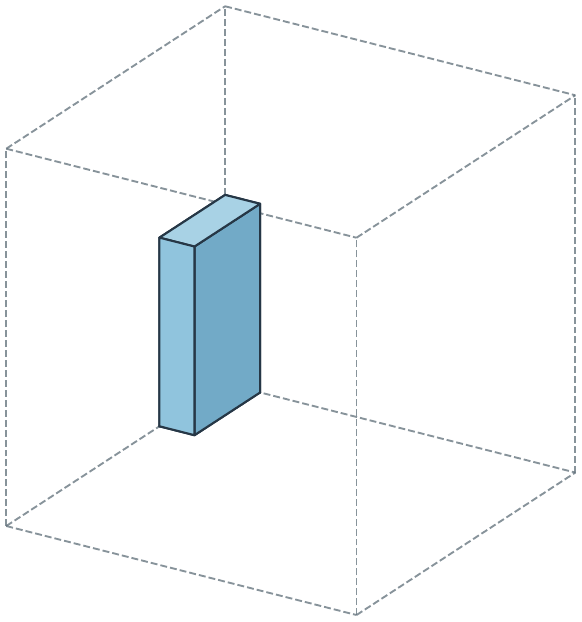}%
    \par\vspace{2pt}
    {\footnotesize \dslabel1-3-5\relax}
  \end{minipage}%
 &
  \begin{minipage}[t][0.85in][t]{0.13\textwidth}
    \centering
    \includegraphics[width=\linewidth,height=0.69in,keepaspectratio]{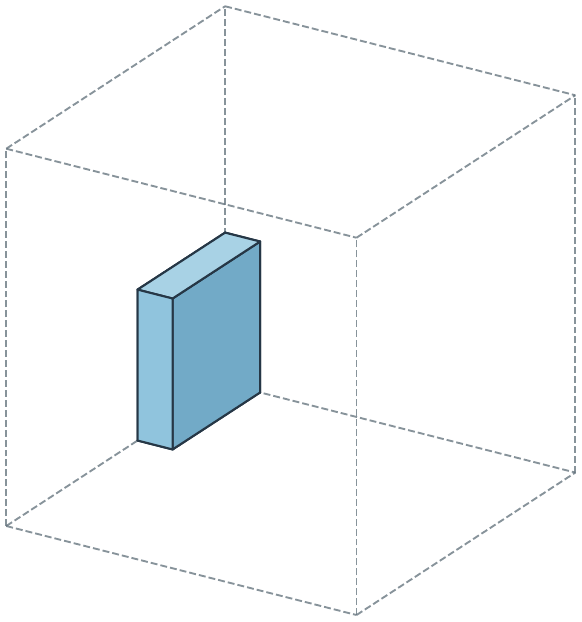}%
    \par\vspace{2pt}
    {\footnotesize \dslabel1-4-4\relax}
  \end{minipage}%
 &
  \begin{minipage}[t][0.85in][t]{0.13\textwidth}
    \centering
    \includegraphics[width=\linewidth,height=0.69in,keepaspectratio]{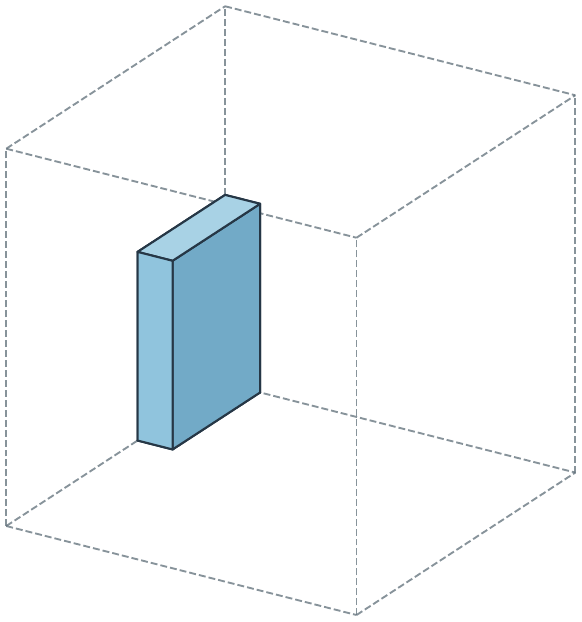}%
    \par\vspace{2pt}
    {\footnotesize \dslabel1-4-5\relax}
  \end{minipage}%
 \\
  \begin{minipage}[t][0.85in][t]{0.13\textwidth}
    \centering
    \includegraphics[width=\linewidth,height=0.69in,keepaspectratio]{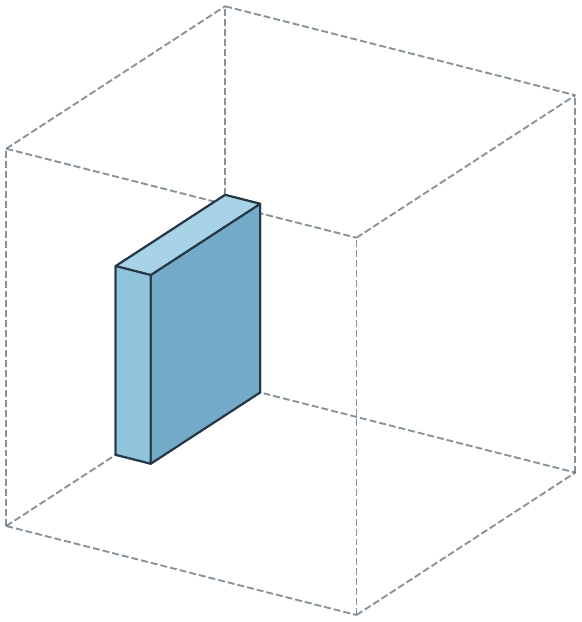}%
    \par\vspace{2pt}
    {\footnotesize \dslabel1-5-5\relax}
  \end{minipage}%
 &
  \begin{minipage}[t][0.85in][t]{0.13\textwidth}
    \centering
    \includegraphics[width=\linewidth,height=0.69in,keepaspectratio]{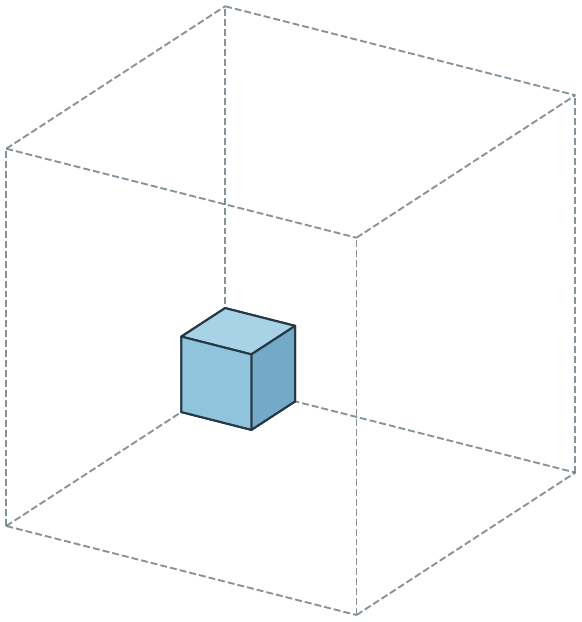}%
    \par\vspace{2pt}
    {\footnotesize \dslabel2-2-2\relax}
  \end{minipage}%
 &
  \begin{minipage}[t][0.85in][t]{0.13\textwidth}
    \centering
    \includegraphics[width=\linewidth,height=0.69in,keepaspectratio]{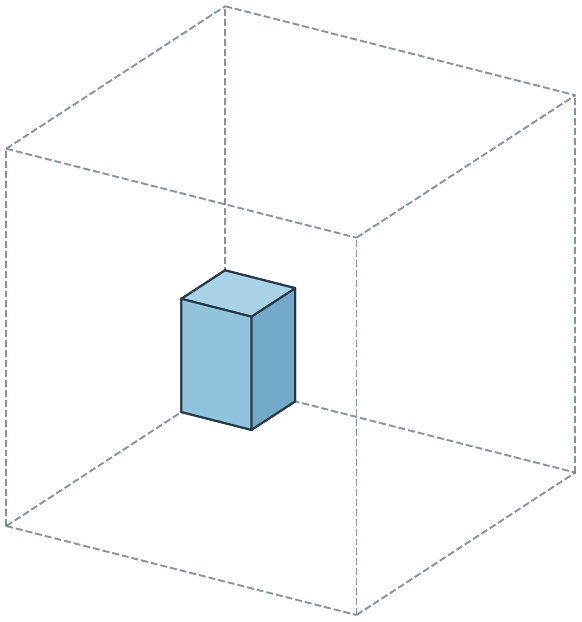}%
    \par\vspace{2pt}
    {\footnotesize \dslabel2-2-3\relax}
  \end{minipage}%
 &
  \begin{minipage}[t][0.85in][t]{0.13\textwidth}
    \centering
    \includegraphics[width=\linewidth,height=0.69in,keepaspectratio]{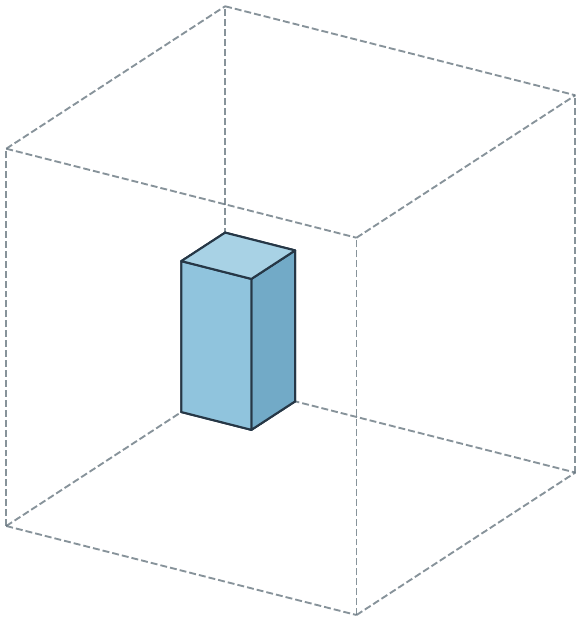}%
    \par\vspace{2pt}
    {\footnotesize \dslabel2-2-4\relax}
  \end{minipage}%
 &
  \begin{minipage}[t][0.85in][t]{0.13\textwidth}
    \centering
    \includegraphics[width=\linewidth,height=0.69in,keepaspectratio]{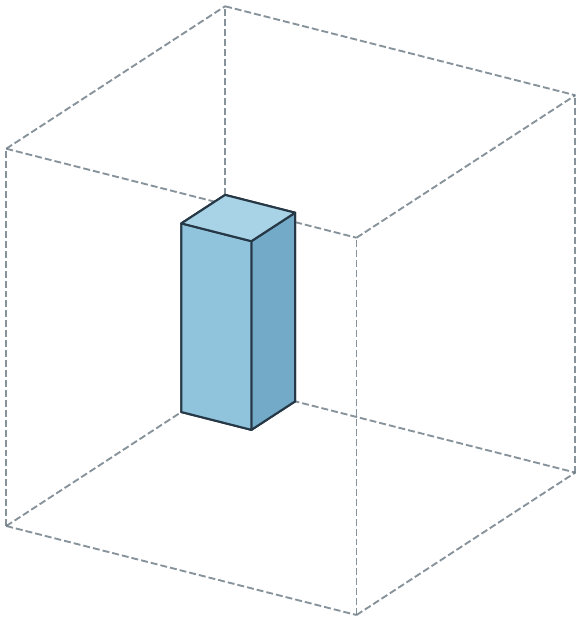}%
    \par\vspace{2pt}
    {\footnotesize \dslabel2-2-5\relax}
  \end{minipage}%
 &
  \begin{minipage}[t][0.85in][t]{0.13\textwidth}
    \centering
    \includegraphics[width=\linewidth,height=0.69in,keepaspectratio]{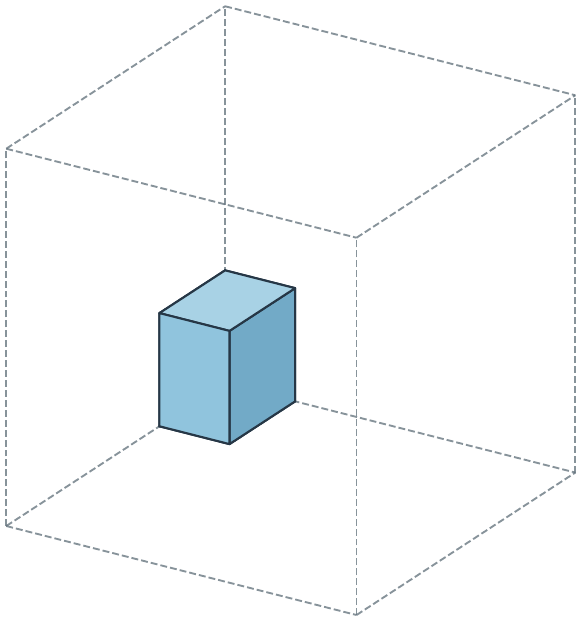}%
    \par\vspace{2pt}
    {\footnotesize \dslabel2-3-3\relax}
  \end{minipage}%
 &
  \begin{minipage}[t][0.85in][t]{0.13\textwidth}
    \centering
    \includegraphics[width=\linewidth,height=0.69in,keepaspectratio]{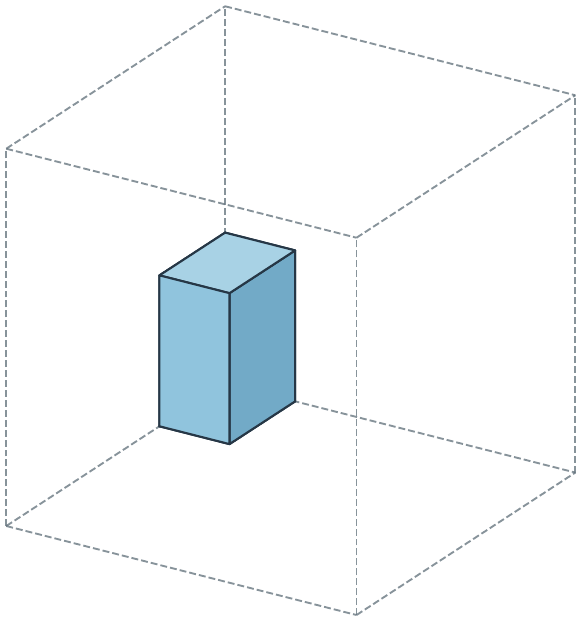}%
    \par\vspace{2pt}
    {\footnotesize \dslabel2-3-4\relax}
  \end{minipage}%
 \\
  \begin{minipage}[t][0.85in][t]{0.13\textwidth}
    \centering
    \includegraphics[width=\linewidth,height=0.69in,keepaspectratio]{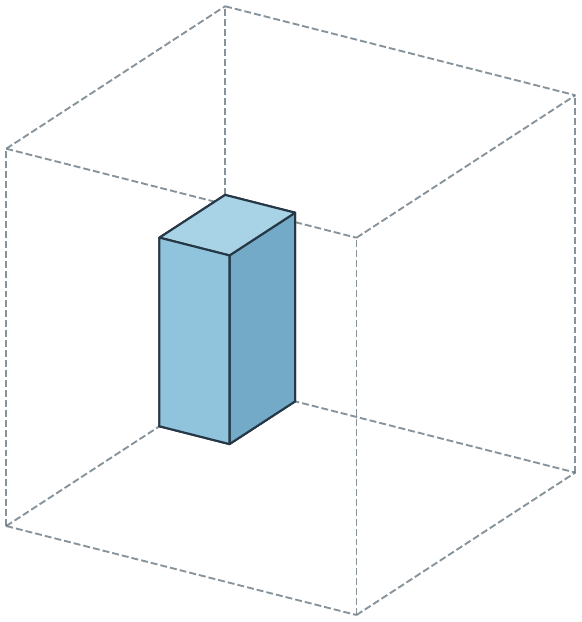}%
    \par\vspace{2pt}
    {\footnotesize \dslabel2-3-5\relax}
  \end{minipage}%
 &
  \begin{minipage}[t][0.85in][t]{0.13\textwidth}
    \centering
    \includegraphics[width=\linewidth,height=0.69in,keepaspectratio]{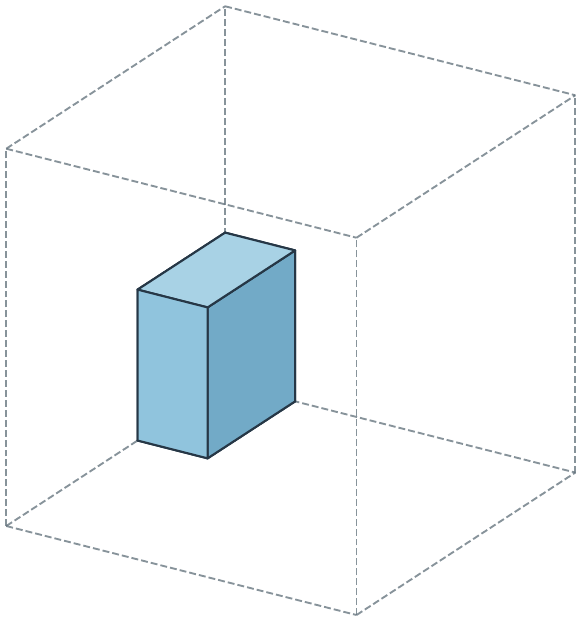}%
    \par\vspace{2pt}
    {\footnotesize \dslabel2-4-4\relax}
  \end{minipage}%
 &
  \begin{minipage}[t][0.85in][t]{0.13\textwidth}
    \centering
    \includegraphics[width=\linewidth,height=0.69in,keepaspectratio]{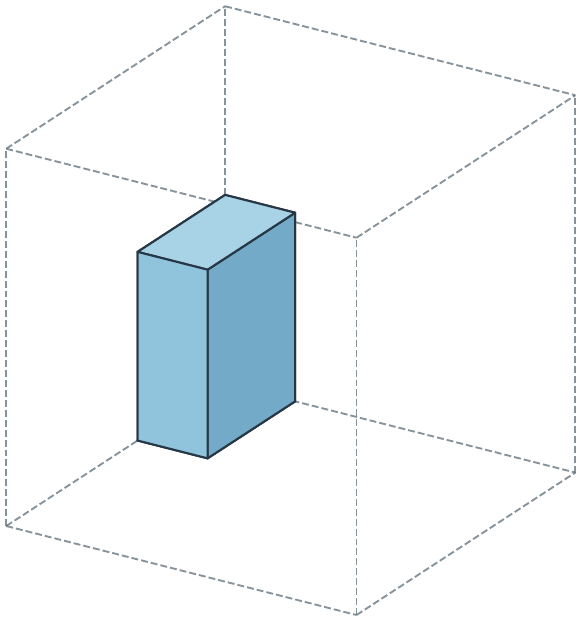}%
    \par\vspace{2pt}
    {\footnotesize \dslabel2-4-5\relax}
  \end{minipage}%
 &
  \begin{minipage}[t][0.85in][t]{0.13\textwidth}
    \centering
    \includegraphics[width=\linewidth,height=0.69in,keepaspectratio]{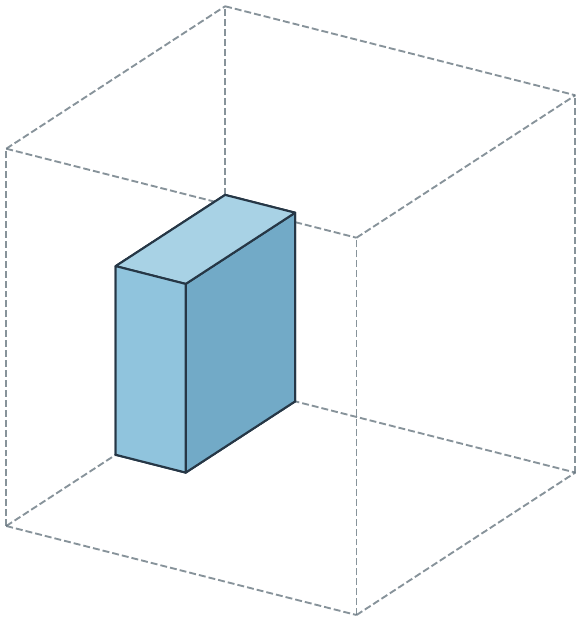}%
    \par\vspace{2pt}
    {\footnotesize \dslabel2-5-5\relax}
  \end{minipage}%
 &
  \begin{minipage}[t][0.85in][t]{0.13\textwidth}
    \centering
    \includegraphics[width=\linewidth,height=0.69in,keepaspectratio]{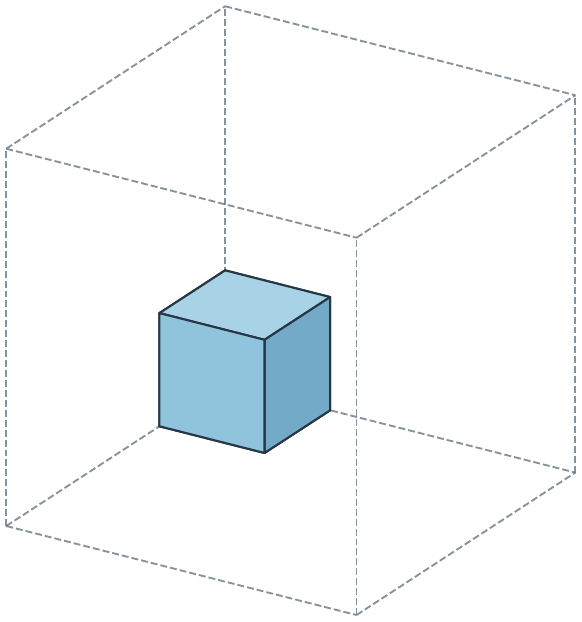}%
    \par\vspace{2pt}
    {\footnotesize \dslabel3-3-3\relax}
  \end{minipage}%
 &
  \begin{minipage}[t][0.85in][t]{0.13\textwidth}
    \centering
    \includegraphics[width=\linewidth,height=0.69in,keepaspectratio]{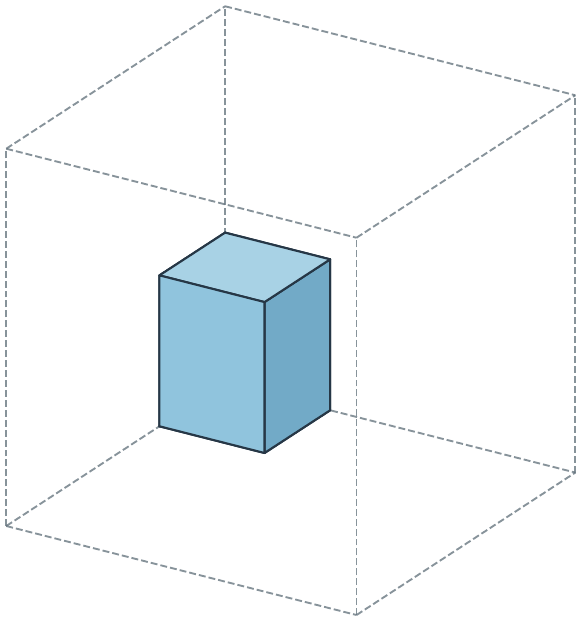}%
    \par\vspace{2pt}
    {\footnotesize \dslabel3-3-4\relax}
  \end{minipage}%
 &
  \begin{minipage}[t][0.85in][t]{0.13\textwidth}
    \centering
    \includegraphics[width=\linewidth,height=0.69in,keepaspectratio]{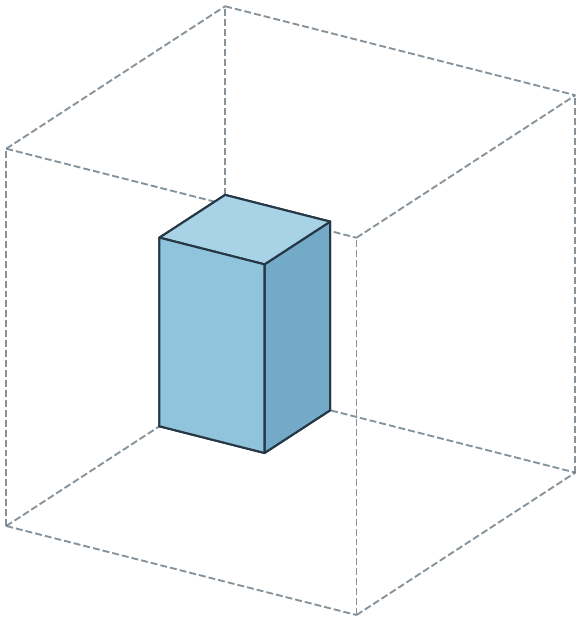}%
    \par\vspace{2pt}
    {\footnotesize \dslabel3-3-5\relax}
  \end{minipage}%
 \\
  \begin{minipage}[t][0.85in][t]{0.13\textwidth}
    \centering
    \includegraphics[width=\linewidth,height=0.69in,keepaspectratio]{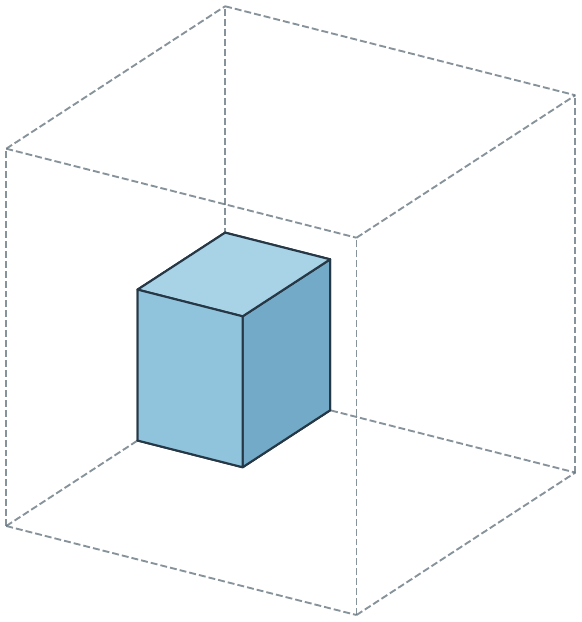}%
    \par\vspace{2pt}
    {\footnotesize \dslabel3-4-4\relax}
  \end{minipage}%
 &
  \begin{minipage}[t][0.85in][t]{0.13\textwidth}
    \centering
    \includegraphics[width=\linewidth,height=0.69in,keepaspectratio]{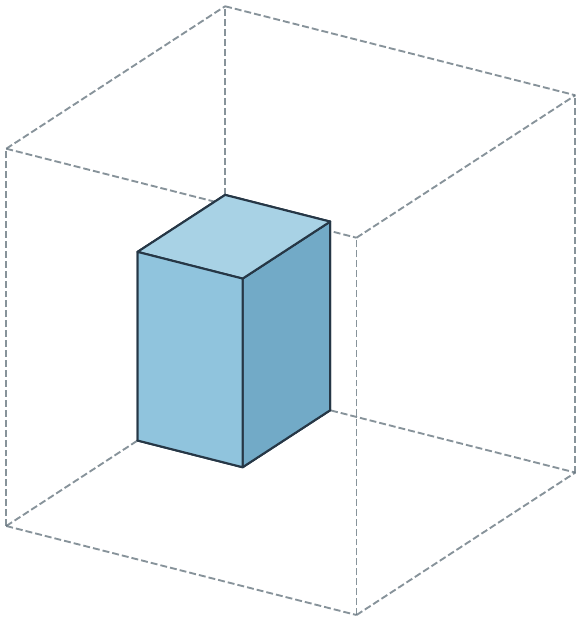}%
    \par\vspace{2pt}
    {\footnotesize \dslabel3-4-5\relax}
  \end{minipage}%
 &
  \begin{minipage}[t][0.85in][t]{0.13\textwidth}
    \centering
    \includegraphics[width=\linewidth,height=0.69in,keepaspectratio]{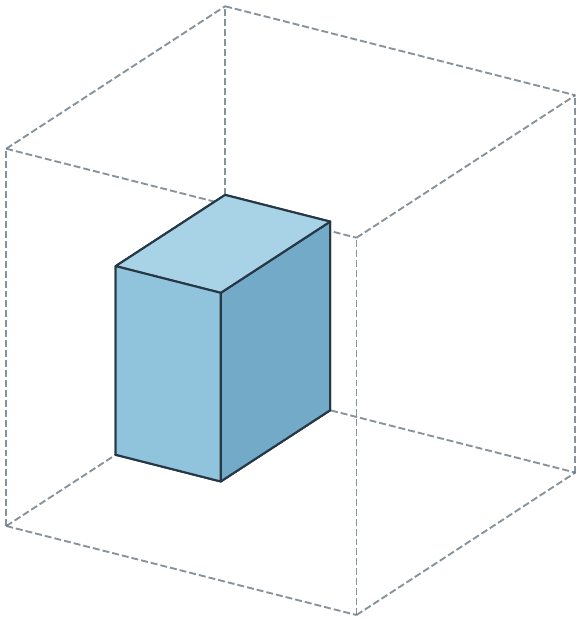}%
    \par\vspace{2pt}
    {\footnotesize \dslabel3-5-5\relax}
  \end{minipage}%
 &
  \begin{minipage}[t][0.85in][t]{0.13\textwidth}
    \centering
    \includegraphics[width=\linewidth,height=0.69in,keepaspectratio]{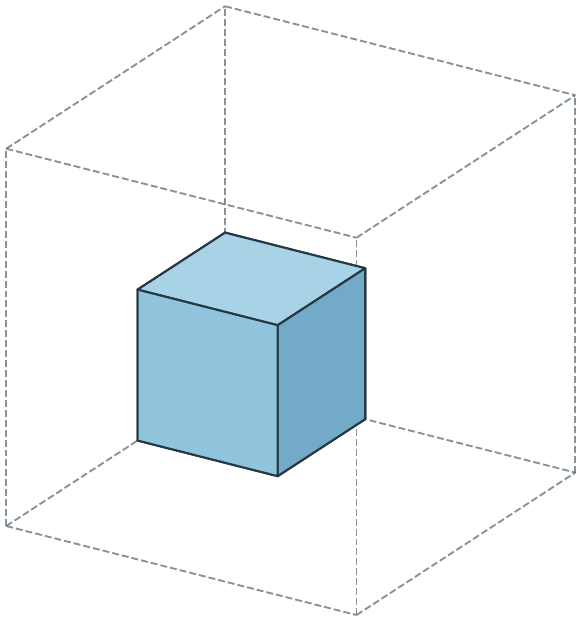}%
    \par\vspace{2pt}
    {\footnotesize \dslabel4-4-4\relax}
  \end{minipage}%
 &
  \begin{minipage}[t][0.85in][t]{0.13\textwidth}
    \centering
    \includegraphics[width=\linewidth,height=0.69in,keepaspectratio]{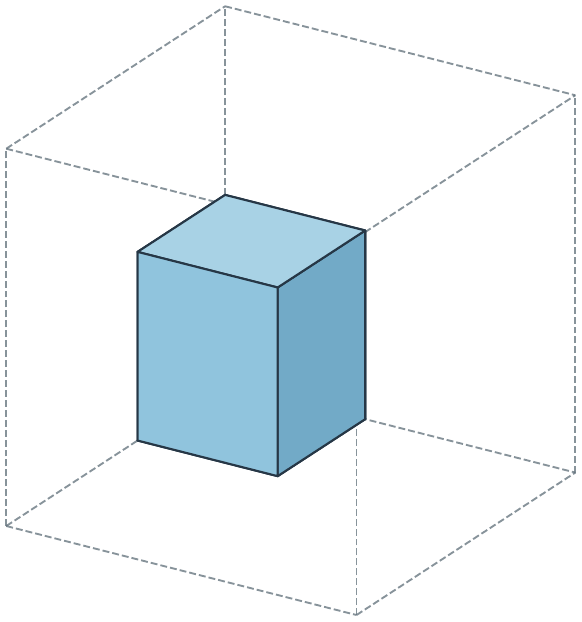}%
    \par\vspace{2pt}
    {\footnotesize \dslabel4-4-5\relax}
  \end{minipage}%
 &
  \begin{minipage}[t][0.85in][t]{0.13\textwidth}
    \centering
    \includegraphics[width=\linewidth,height=0.69in,keepaspectratio]{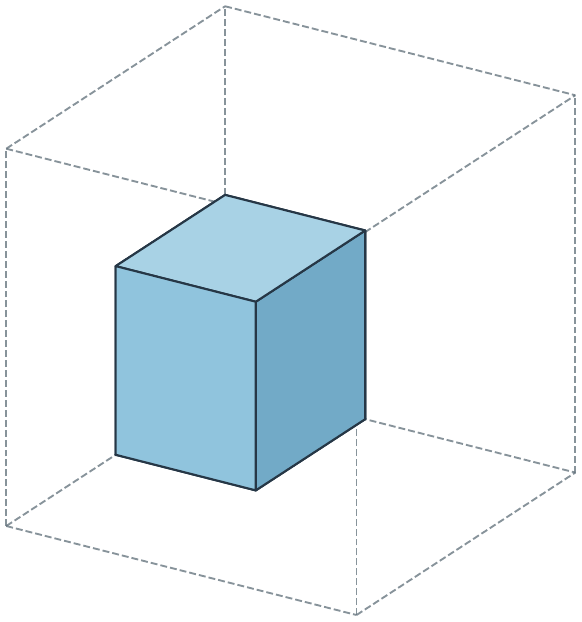}%
    \par\vspace{2pt}
    {\footnotesize \dslabel4-5-5\relax}
  \end{minipage}%
 &
  \begin{minipage}[t][0.85in][t]{0.13\textwidth}
    \centering
    \includegraphics[width=\linewidth,height=0.69in,keepaspectratio]{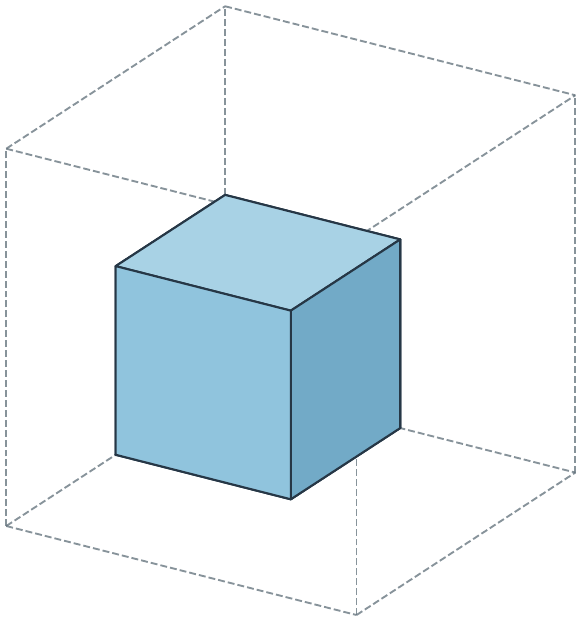}%
    \par\vspace{2pt}
    {\footnotesize \dslabel5-5-5\relax}
  \end{minipage}%

\end{tabular*}
\par
\endgroup
\par\vspace{2pt}
{\normalsize (a) Dataset 1 ($\mathcal{D}_1$)}\par

\vspace{0.18in}

\setlength{\dscolumnwidth}{0.5\textwidth}
\addtolength{\dscolumnwidth}{-0.5\columnsep}
\noindent
\begin{minipage}[t]{\dscolumnwidth}
\centering
\setlength{\tabcolsep}{0pt}
\begin{tabular*}{\linewidth}{@{\extracolsep{\fill}}ccc@{}}
  \begin{minipage}[t][0.94in][t]{0.30\linewidth}
    \centering
    \includegraphics[width=\linewidth,height=0.74in,keepaspectratio]{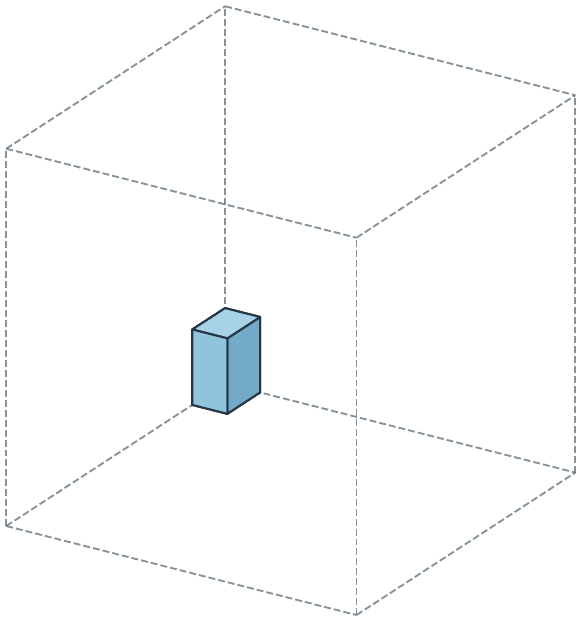}%
    \par\vspace{2pt}
    {\scriptsize \dslabel1-1.5-2\relax}
  \end{minipage}%
 &
  \begin{minipage}[t][0.94in][t]{0.30\linewidth}
    \centering
    \includegraphics[width=\linewidth,height=0.74in,keepaspectratio]{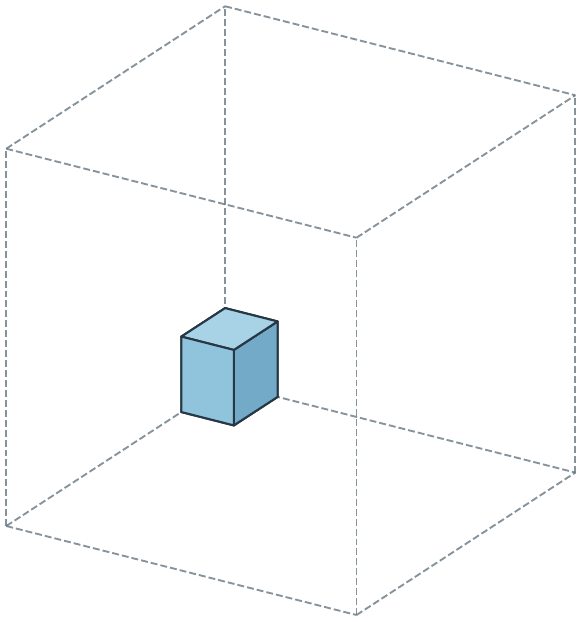}%
    \par\vspace{2pt}
    {\scriptsize \dslabel1.5-2-2\relax}
  \end{minipage}%
 &
  \begin{minipage}[t][0.94in][t]{0.30\linewidth}
    \centering
    \includegraphics[width=\linewidth,height=0.74in,keepaspectratio]{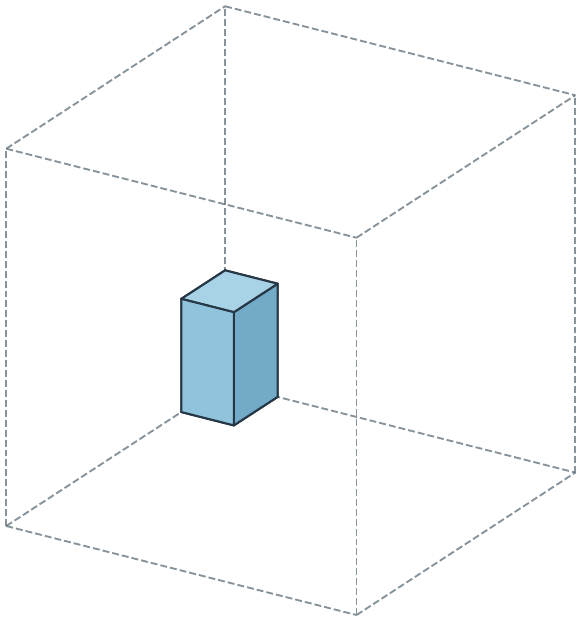}%
    \par\vspace{2pt}
    {\scriptsize \dslabel1.5-2-3\relax}
  \end{minipage}%
 \\
  \begin{minipage}[t][0.94in][t]{0.30\linewidth}
    \centering
    \includegraphics[width=\linewidth,height=0.74in,keepaspectratio]{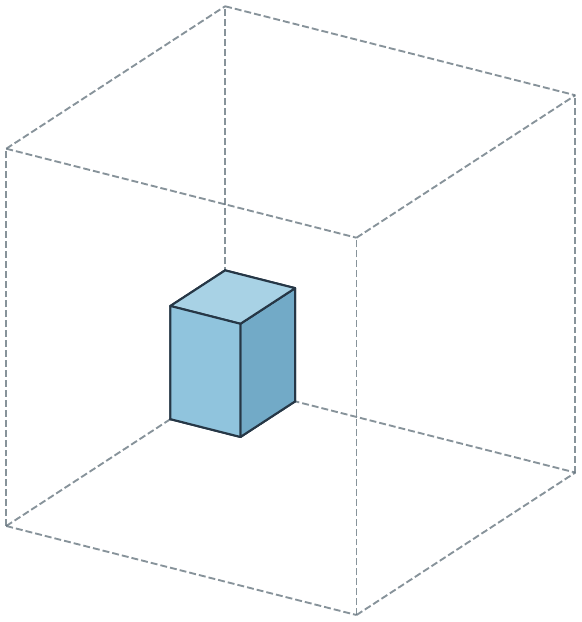}%
    \par\vspace{2pt}
    {\scriptsize \dslabel2-2.5-3\relax}
  \end{minipage}%
 &
  \begin{minipage}[t][0.94in][t]{0.30\linewidth}
    \centering
    \includegraphics[width=\linewidth,height=0.74in,keepaspectratio]{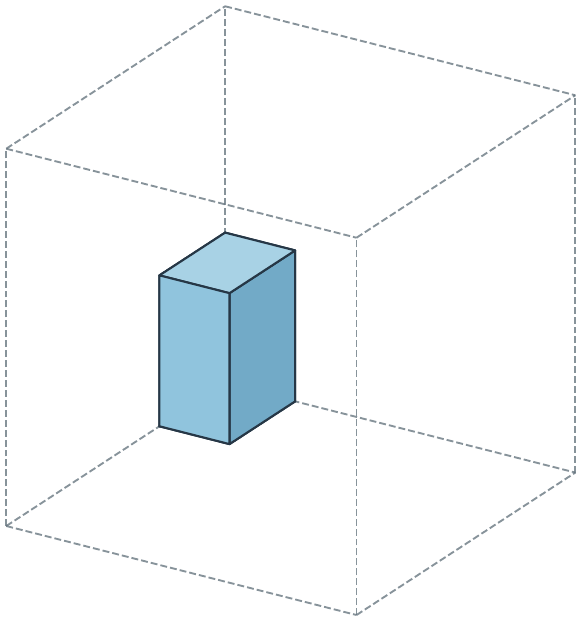}%
    \par\vspace{2pt}
    {\scriptsize \dslabel2-3-4\relax}
  \end{minipage}%
 &
  \begin{minipage}[t][0.94in][t]{0.30\linewidth}
    \centering
    \includegraphics[width=\linewidth,height=0.74in,keepaspectratio]{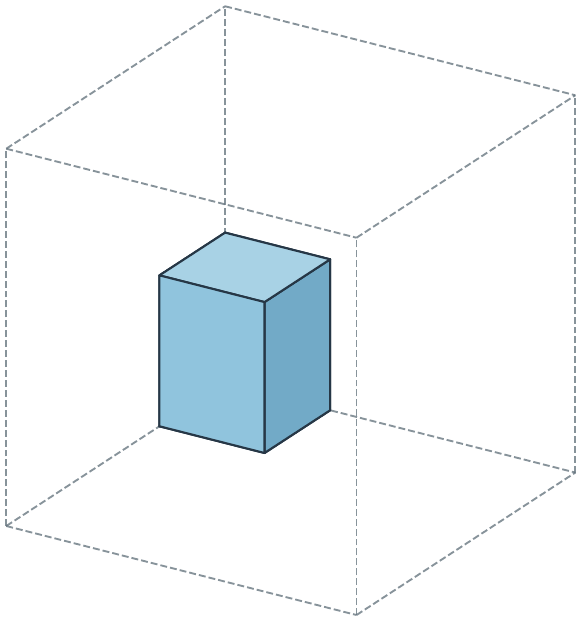}%
    \par\vspace{2pt}
    {\scriptsize \dslabel3-3-4\relax}
  \end{minipage}%
 \\
  \begin{minipage}[t][0.94in][t]{0.30\linewidth}
    \centering
    \includegraphics[width=\linewidth,height=0.74in,keepaspectratio]{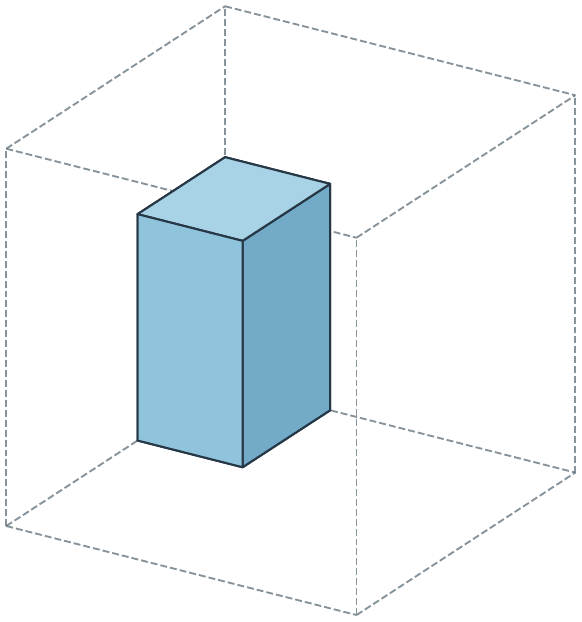}%
    \par\vspace{2pt}
    {\scriptsize \dslabel3-4-6\relax}
  \end{minipage}%
 &
  \begin{minipage}[t][0.94in][t]{0.30\linewidth}
    \centering
    \includegraphics[width=\linewidth,height=0.74in,keepaspectratio]{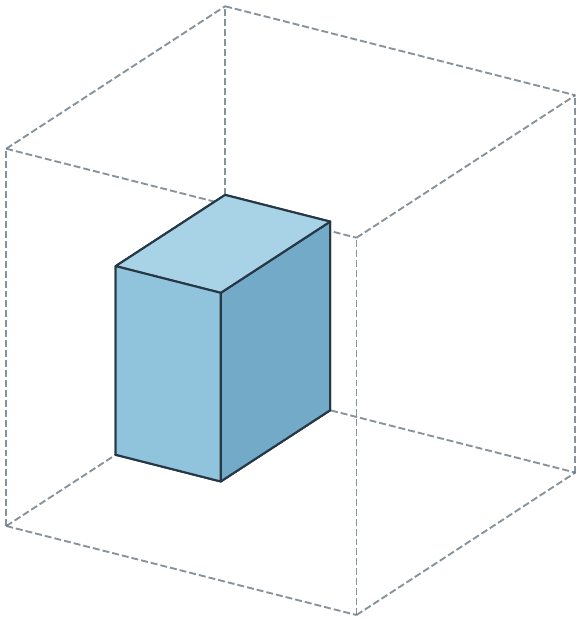}%
    \par\vspace{2pt}
    {\scriptsize \dslabel3-5-5\relax}
  \end{minipage}%
 &
  \begin{minipage}[t][0.94in][t]{0.30\linewidth}
    \centering
    \includegraphics[width=\linewidth,height=0.74in,keepaspectratio]{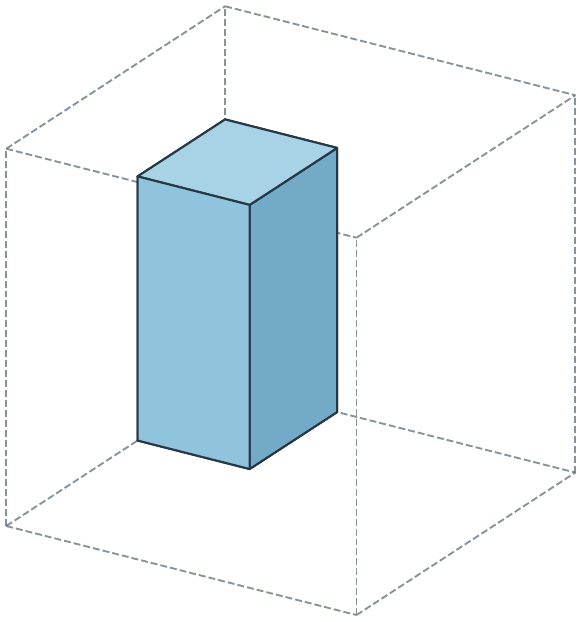}%
    \par\vspace{2pt}
    {\scriptsize \dslabel3.2-4-7\relax}
  \end{minipage}%
 \\
  \begin{minipage}[t][0.94in][t]{0.30\linewidth}
    \centering
    \includegraphics[width=\linewidth,height=0.74in,keepaspectratio]{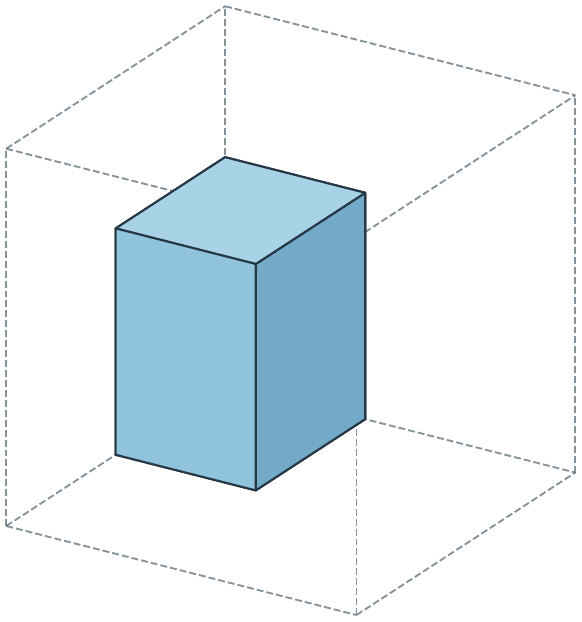}%
    \par\vspace{2pt}
    {\scriptsize \dslabel4-5-6\relax}
  \end{minipage}%
 & &
\end{tabular*}
\par\vspace{2pt}
{\normalsize (b) Dataset 2 ($\mathcal{D}_2$)}
\end{minipage}\hspace{\columnsep}%
\begin{minipage}[t]{\dscolumnwidth}
\centering
\setlength{\tabcolsep}{0pt}
\begin{tabular*}{\linewidth}{@{\extracolsep{\fill}}ccc@{}}
  \begin{minipage}[t][0.94in][t]{0.30\linewidth}
    \centering
    \includegraphics[width=\linewidth,height=0.74in,keepaspectratio]{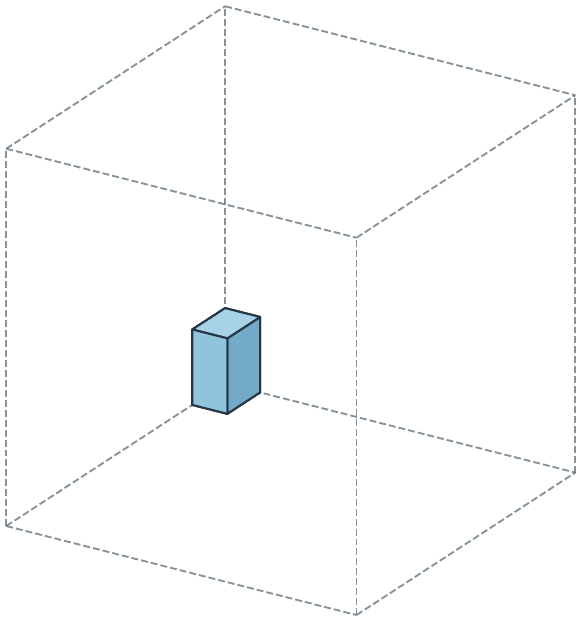}%
    \par\vspace{2pt}
    {\scriptsize \dslabel1-1.5-2\relax}
  \end{minipage}%
 &
  \begin{minipage}[t][0.94in][t]{0.30\linewidth}
    \centering
    \includegraphics[width=\linewidth,height=0.74in,keepaspectratio]{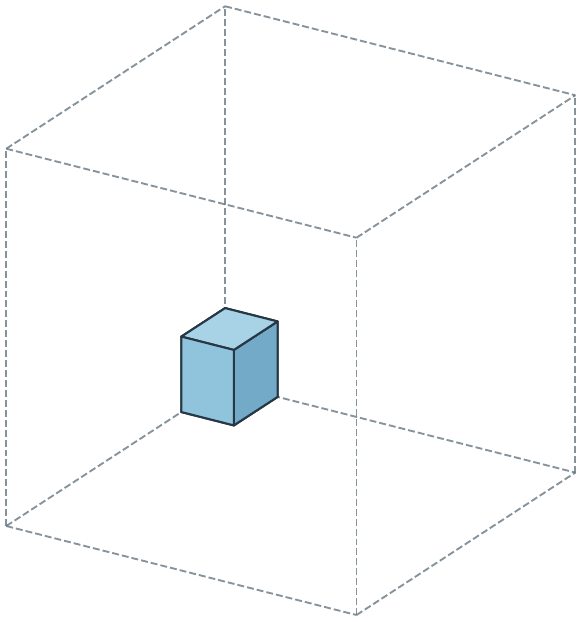}%
    \par\vspace{2pt}
    {\scriptsize \dslabel1.5-2-2\relax}
  \end{minipage}%
 &
  \begin{minipage}[t][0.94in][t]{0.30\linewidth}
    \centering
    \includegraphics[width=\linewidth,height=0.74in,keepaspectratio]{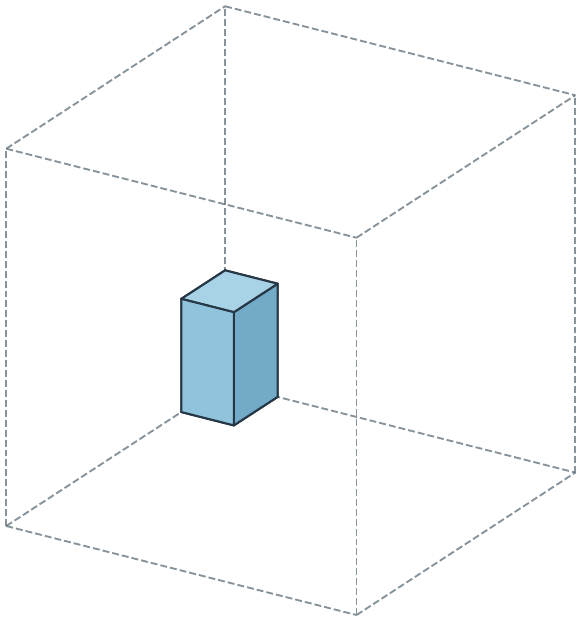}%
    \par\vspace{2pt}
    {\scriptsize \dslabel1.5-2-3\relax}
  \end{minipage}%
 \\
  \begin{minipage}[t][0.94in][t]{0.30\linewidth}
    \centering
    \includegraphics[width=\linewidth,height=0.74in,keepaspectratio]{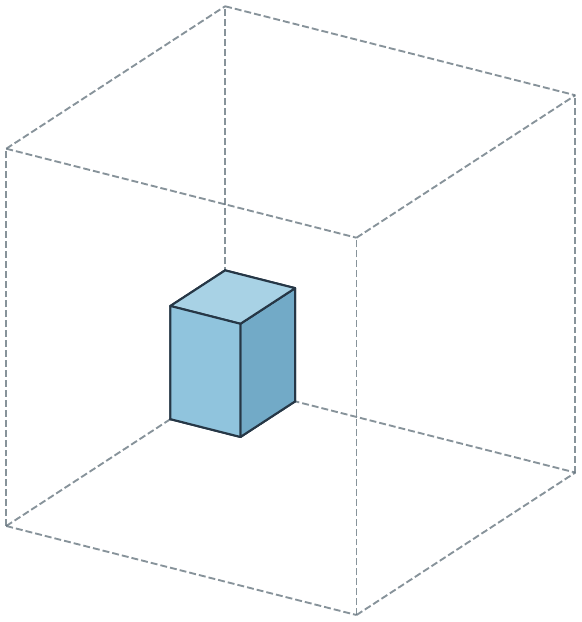}%
    \par\vspace{2pt}
    {\scriptsize \dslabel2-2.5-3\relax}
  \end{minipage}%
 &
  \begin{minipage}[t][0.94in][t]{0.30\linewidth}
    \centering
    \includegraphics[width=\linewidth,height=0.74in,keepaspectratio]{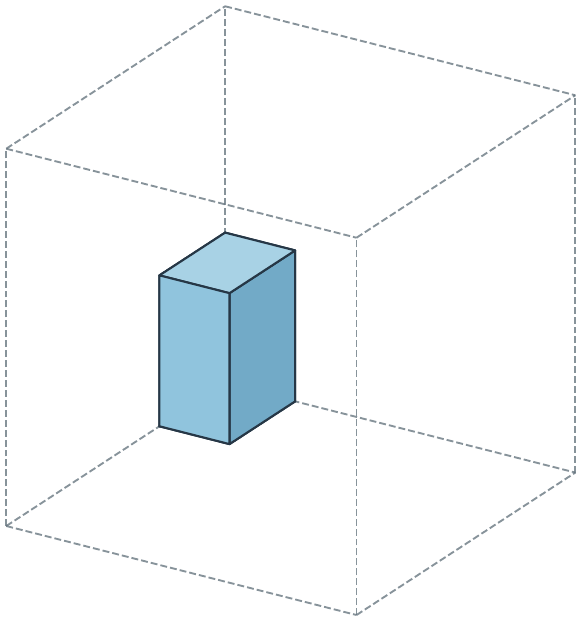}%
    \par\vspace{2pt}
    {\scriptsize \dslabel2-3-4\relax}
  \end{minipage}%
 &
  \begin{minipage}[t][0.94in][t]{0.30\linewidth}
    \centering
    \includegraphics[width=\linewidth,height=0.74in,keepaspectratio]{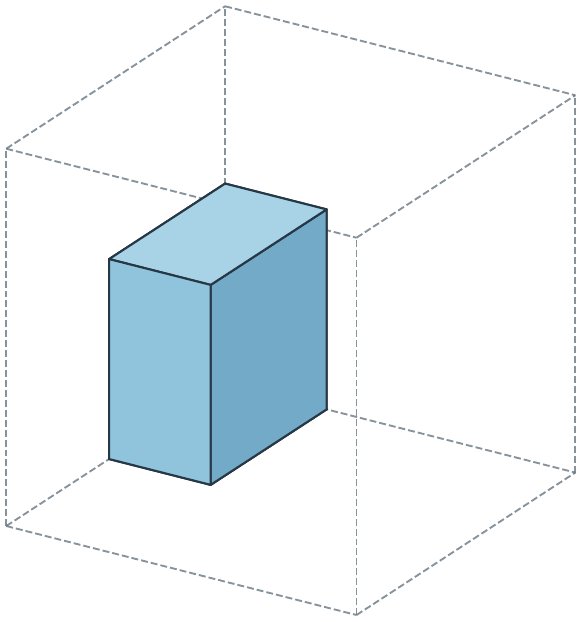}%
    \par\vspace{2pt}
    {\scriptsize \dslabel2.9-5.3-5.3\relax}
  \end{minipage}%
 \\
  \begin{minipage}[t][0.94in][t]{0.30\linewidth}
    \centering
    \includegraphics[width=\linewidth,height=0.74in,keepaspectratio]{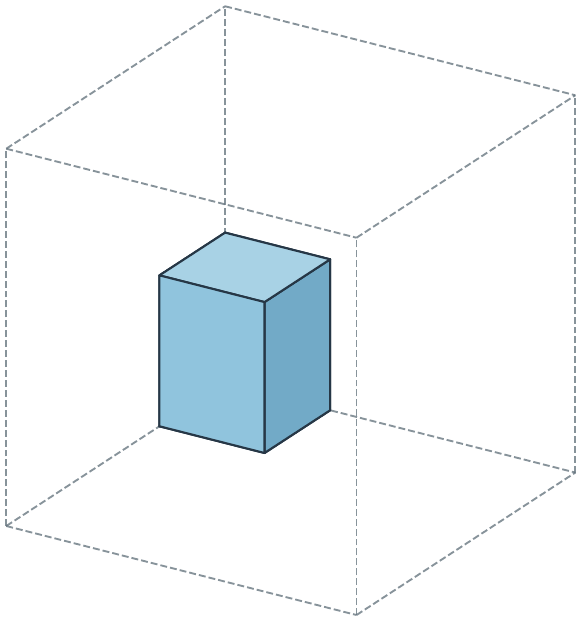}%
    \par\vspace{2pt}
    {\scriptsize \dslabel3-3-4\relax}
  \end{minipage}%
 &
  \begin{minipage}[t][0.94in][t]{0.30\linewidth}
    \centering
    \includegraphics[width=\linewidth,height=0.74in,keepaspectratio]{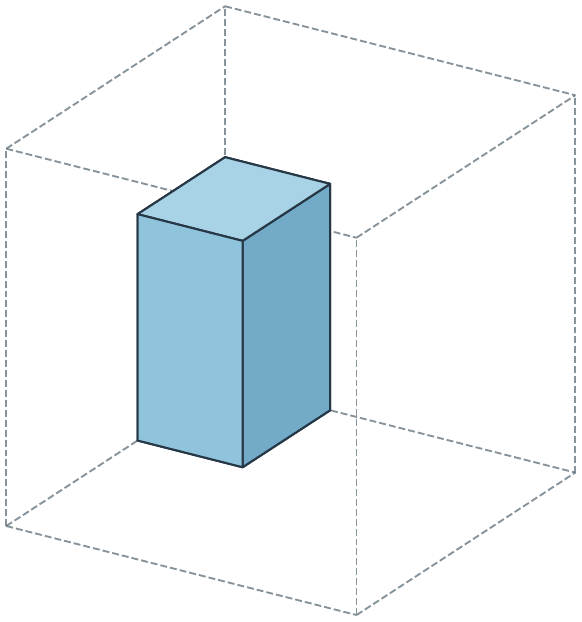}%
    \par\vspace{2pt}
    {\scriptsize \dslabel3-4-6\relax}
  \end{minipage}%
 &
  \begin{minipage}[t][0.94in][t]{0.30\linewidth}
    \centering
    \includegraphics[width=\linewidth,height=0.74in,keepaspectratio]{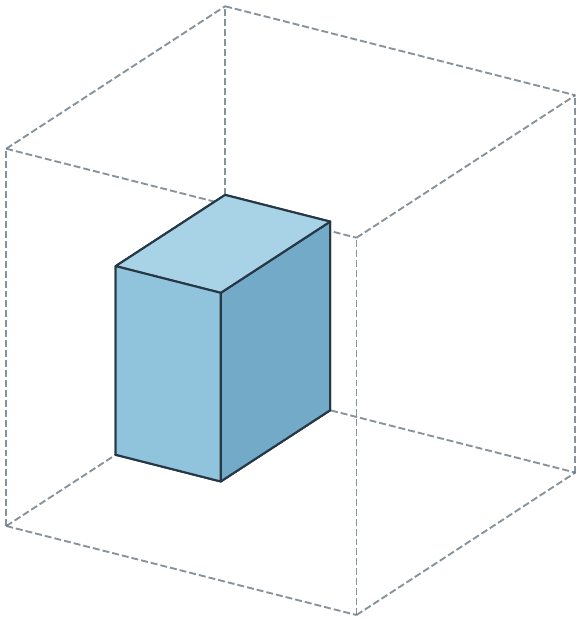}%
    \par\vspace{2pt}
    {\scriptsize \dslabel3-5-5\relax}
  \end{minipage}%
 \\
  \begin{minipage}[t][0.94in][t]{0.30\linewidth}
    \centering
    \includegraphics[width=\linewidth,height=0.74in,keepaspectratio]{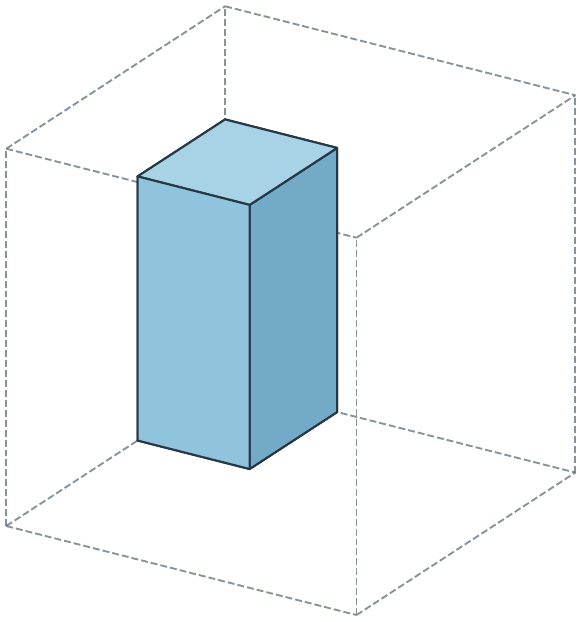}%
    \par\vspace{2pt}
    {\scriptsize \dslabel3.2-4-7\relax}
  \end{minipage}%
 &
  \begin{minipage}[t][0.94in][t]{0.30\linewidth}
    \centering
    \includegraphics[width=\linewidth,height=0.74in,keepaspectratio]{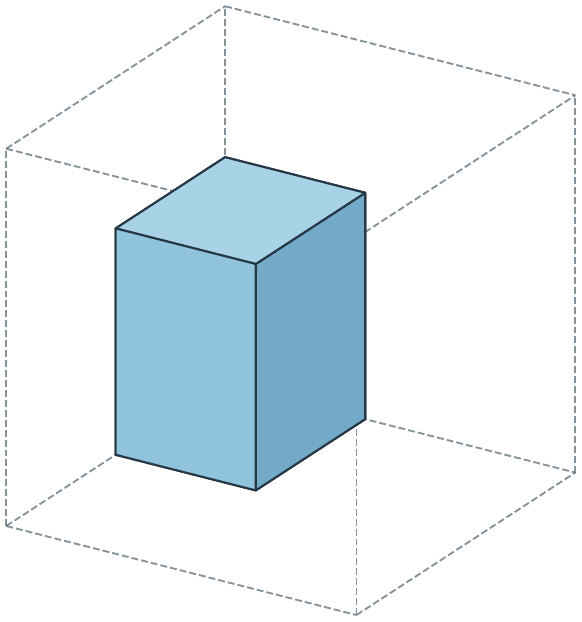}%
    \par\vspace{2pt}
    {\scriptsize \dslabel4-5-6\relax}
  \end{minipage}%
 &
\end{tabular*}
\par\vspace{2pt}
{\normalsize (c) Dataset 3 ($\mathcal{D}_3$)}
\end{minipage}

\caption{Item types in the three datasets. Each cube is rendered at the relative dimensions of one item type within a unit container and labeled by its dimensions in ascending order. (a)~Dataset~1 ($\mathcal{D}_1$), (b)~Dataset~2 ($\mathcal{D}_2$), and (c)~Dataset~3 ($\mathcal{D}_3$).}
\label{fig:dataset-visualization}
\end{figure*}

\twocolumn
\printcredits

\end{document}